\documentclass{article}

\usepackage[final]{neurips_2024}

\usepackage[utf8]{inputenc} %
\usepackage[T1]{fontenc}    %
\usepackage{hyperref}       %
\hypersetup{
    colorlinks=true,
    anchorcolor=blue,
    citecolor={blue!60!black},
    linkcolor=blue
}
\usepackage{url}            %
\usepackage{booktabs}       %
\usepackage{amsfonts}       %
\usepackage{nicefrac}       %
\usepackage{microtype}      %
\usepackage{xcolor}         %
\usepackage{xspace}
\usepackage{colortbl}
\usepackage{amsmath}
\usepackage{graphicx}
\usepackage{appendix}
\usepackage[algo2e]{algorithm2e} 
\usepackage{algorithm,algorithmic}
\usepackage{multirow}
\usepackage{wrapfig}
\usepackage{subcaption}
\usepackage{enumitem}

\DeclareMathAlphabet\mathbfcal{OMS}{cmsy}{b}{n}
\makeatletter
\newcommand{\minisection}[1]{\vspace{1pt}\noindent\textbf{#1}}
\newcommand{\xRightarrow}[2][]{\ext@arrow 0359\Rightarrowfill@{#1}{#2}}
\makeatother

\newcommand{\fig}[1]{Figure~\ref{#1}}
\newcommand{\eq}[1]{Equation~(\ref{#1})}
\newcommand{\tb}[1]{Table~\ref{#1}}
\newcommand{\se}[1]{Section~\ref{#1}}

\newcommand{\bbE}{\ensuremath{\mathbb{E}}} %
\newcommand{\caN}{\ensuremath{\mathcal{N}}} %
\newcommand{\caL}{\ensuremath{\mathcal{L}}} %
\newcommand{\caD}{\ensuremath{\mathcal{D}}} %
\newcommand{\caU}{\ensuremath{\mathcal{U}}} %

\usepackage{amsmath,amsfonts,bm}

\def\Secref#1{Section~\ref{#1}}

\def\eqref#1{equation~\ref{#1}}

\def\1{\bm{1}}

\DeclareMathAlphabet{\mathsfit}{\encodingdefault}{\sfdefault}{m}{sl}
\SetMathAlphabet{\mathsfit}{bold}{\encodingdefault}{\sfdefault}{bx}{n}

\def\gN{{\mathcal{N}}}

\newcommand{\our}{\textsc{MADiff}\xspace}

\newcommand{\blue}[1]{\textcolor{blue}{#1}}

\newcolumntype{a}{>{\columncolor{red!15}}c}

\title{\our: Offline Multi-agent Learning\\ with Diffusion Models}

\makeatletter
\def\@fnsymbol#1{\ifcase#1\or \dag\else\@ctrerr\fi}
\makeatother

\author{
   Zhengbang Zhu$^{1}$~~~ Minghuan Liu$^{1}$~~~ Liyuan Mao$^{1}$~~~ Bingyi Kang$^{2}$~~~ Minkai Xu$^{3}$\\
   \textbf{Yong Yu$^{1}$~~~ Stefano Ermon$^{3}$~~~ Weinan Zhang$^{1}$\thanks{Corresponding author.}}\\
  {$^1$ Shanghai Jiao Tong University, $^2$ ByteDance, $^3$ Stanford University}\\
  \texttt{\{zhengbangzhu, minghuanliu, maoliyuan, yyu, wnzhang\}@sjtu.edu.cn}, \\
  \texttt{bingykang@gmail.com}, \texttt{\{minkai, ermon\}@cs.stanford.edu}
}

\begin{document}

\maketitle

\begin{abstract}
Offline reinforcement learning (RL) aims to learn policies from pre-existing datasets without further interactions, making it a challenging task. Q-learning algorithms struggle with extrapolation errors in offline settings, while supervised learning methods are constrained by model expressiveness. Recently, diffusion models (DMs) have shown promise in overcoming these limitations in single-agent learning, but their application in multi-agent scenarios remains unclear. Generating trajectories for each agent with independent DMs may impede coordination, while concatenating all agents' information can lead to low sample efficiency. Accordingly, we propose \our, which is realized with an attention-based diffusion model to model the complex coordination among behaviors of multiple agents. To our knowledge, \our is the first diffusion-based multi-agent learning framework, functioning as both a decentralized policy and a centralized controller. During decentralized executions, \our simultaneously performs teammate modeling, and the centralized controller can also be applied in multi-agent trajectory predictions. Our experiments demonstrate that \our outperforms baseline algorithms across various multi-agent learning tasks, highlighting its effectiveness in modeling complex multi-agent interactions.
\end{abstract}

\section{Introduction}
Offline reinforcement learning (RL)~\citep{fujimoto2019off,kumar2020conservative} learns exclusively from static datasets without online interactions, enabling the effective use of pre-collected large-scale data.
However, applying temporal difference (TD) learning in offline settings causes extrapolation errors~\citep{fujimoto2019off}, where target value functions are evaluated on out-of-distribution actions. 
Sequence modeling algorithms bypass TD-learning by directly fitting the dataset distribution~\citep{chen2021decision, janner2021offline}.
Nevertheless, these methods are limited by the model's expressiveness, making it difficult to handle diverse datasets. They also suffer from compounding errors~\citep{xiao2019learning} due to autoregressive generation.
Recently, diffusion models (DMs) have achieved remarkable success in various generative modeling tasks~\citep{song2019generative,ho2020denoising,xu2022geodiff}, owing to their exceptional abilities at capturing complex, high-dimensional data distributions.
Their successes have also been introduced into offline RL, offering a superior modeling choice for sequence modeling algorithms~\citep{janner2022planning,ajay2022conditional}.

Compared to single-agent learning, offline multi-agent learning (MAL) has been less studied and is more challenging.
Since the behaviors of all agents are interrelated, each agent is required to model interactions and coordination among agents, while making decisions in a decentralized manner to achieve the goal.
Current MAL approaches typically train a centralized value function to update individual agents' policies~\citep{rashid2020monotonic} or use an autoregressive transformer to determine each agent's actions~\citep{meng2021offline,wen2022multi}. However, without online interactions, an incorrect centralized value can lead to significant extrapolation errors, and the transformer can only serve as an independent model for each agent.

In this paper, we aim to study the potential of employing DMs to solve the above challenges in offline MAL problems. Merely adopting existing diffusion RL methods by using independent DMs to model each agent can result in serious inconsistencies due to a lack of proper credit assignment among agents. 
Another possible solution is to concatenate all agents' information as the input and output of the DM. However, treating the agents as a single unified agent neglects the important nature of multi-agent systems. 
One agent may have strong correlations with only a few other agents, which makes a full feature interaction redundant.
In many multi-agent systems, agents exhibit certain symmetry and can share model parameters for efficient learning~\citep{arel2010reinforcement}.
However, concatenating them in a fixed order breaks this symmetry, forcing the model to treat each agent differently.

To address the aforementioned coordination challenges, we propose the first centralized-training-decentralized-execution (CTDE) diffusion framework for MA problems, named \our. \our adopts a novel attention-based DM to learn a return-conditional trajectory generation model on a reward-labeled multi-agent interaction dataset.
In particular, the designed attention is computed in several latent layers of the model of each agent to fully interchange the information and integrate the global information of all agents.
To model the coordination among agents, \our applies the attention mechanism on latent embedding for information interaction across agents.
The attention mechanism enables the dynamic modeling of agent interactions through learned weights, while also enabling the use of a shared backbone to model each agent's trajectory, significantly reducing the number of parameters.
During training, \our performs centralized training on the joint trajectory distributions of all agents from offline datasets, including different levels of expected returns.
During inference, \our adopts classifier-free guidance with low-temperature sampling to generate behaviors given the conditioned high expected returns, allowing for decentralized execution by predicting the behavior of other agents and generating its own behavior.
Therefore, \our can be regarded as a principled offline MAL solution that not only serves as a decentralized policy for each agent or a centralized controller for all agents, but also includes teammate modeling without additional cost.
Comprehensive experiments demonstrated superior performances of \our on various multi-agent learning tasks, including offline MARL and trajectory prediction.

In summary, our contributions are
(1) the first diffusion-based multi-agent learning framework
that unifies decentralized policy, centralized controller, teammate modeling, and trajectory prediction;
(2) a novel attention-based DM structure that is designed explicitly for MAL and enables coordination among agents in each denoising step;
(3) achieving superior performances for various offline multi-agent problems.

\section{Preliminaries}

\subsection{Multi-agent Offline Reinforcement Learning}
We consider a partially observable and fully cooperative multi-agent learning (MAL) problem, where agents with local observations cooperate to finish the task. Formally, it is defined as a Dec-POMDP~\citep{oliehoek2016concise}: $G=\langle \mathcal{S}, \mathcal{A}, P, r, \Omega, O, N, U,\gamma\rangle$, where $\mathcal{S}$ and $\mathcal{A}$ denote state and action space separately, and $\gamma$ is the discounted factor.
The system includes $N$ agents $\{1, 2, \dots, N\}$ act in discrete time steps, and starts with an initial global state $s_0\in\mathcal{S}$ sampled from the distribution $U$.
At each time step $t$, every agent $i$ only observes a local observation $o^i\in\Omega$ produced by the function $O(s, a): \mathcal{S}\times \mathcal{A}\rightarrow\Omega$ and decides $a\in\mathcal{A}$, which forms the joint action $\mathbf{a}\in\mathbfcal{A}\equiv \mathcal{A}^N$, leading the system transits to the next state $s'$ according to the dynamics function $P(s'|s, \mathbf{a}): \mathcal{S}\times\mathbfcal{A}\rightarrow\mathcal{S}$. Normally, agents receive a shared reward $r(s, \mathbf{a})$ at each step, and the optimization objective is to learn a policy $\pi^i$ for each agent that maximizes the discounted cumulative reward $\mathbb{E}_{s_t,\mathbf{a}_t}[\sum_t \gamma^t r(s_t, \mathbf{a}_t)]$. 
In offline settings, instead of collecting online data in environments, we only have access to a static dataset $\mathcal{D}$ to learn the policies. The dataset $\mathcal{D}$ is generally composed of trajectories $\bm{\tau}$, \textit{i.e.}, observation-action sequences $[\bm{o}_0,\bm{a}_0,\bm{o}_1,\bm{a}_{1},\cdots,\bm{o}_{T},\bm{a}_{T}]$ or observation sequences $[\bm{o}_0,\bm{o}_{1},\cdots,\bm{o}_{T}]$. 
We use bold symbols to denote the joint vectors of all agents.

\subsection{Diffusion Probabilistic Models}
Diffusion models (DMs)~\citep{sohl2015deep,song2019generative,ho2020denoising}, as a powerful class of generative models, implement the data generation process as reversing a forward noising process (denoising process). For each data point $x_0\sim p_{\text{data}}(x)$ from the dataset $\caD$, the noising process is a discrete Markov chain $x_{0:K}$ such that $p(x_k|x_{k-1})=\mathcal{N}(x_k|\sqrt{\alpha_k}x_{k-1}, (1-\alpha_k)I)$, where $\mathcal{N}(\mu, \Sigma)$ denotes a Gaussian distribution with mean $\mu$ and variance $\Sigma$, and $\alpha_{0:K}\in\mathbb{R}$ are hyperparameters which control the variance schedule. The variational reverse Markov chain is parameterized with $q_\theta(x_{k-1}|x_k)=\mathcal{N}(x_{k-1}|\mu_{\theta}(x_k, k), (1-\alpha_k)I)$. The data sampling process begins by sampling an initial noise $x_K\sim\mathcal{N}(0,I)$, and follows the reverse process until $x_0$. The reverse process can be estimated by optimizing a simplified surrogate loss as in \cite{ho2020denoising}:
\begin{equation}
    \mathcal{L}(\theta)=\mathbb{E}_{k \sim[1, K], x_0 \sim q, \epsilon \sim \mathcal{N}(0, I)}\left[\left\|\epsilon-\epsilon_\theta\left(x_k, k\right)\right\|^2\right]~.
\end{equation}
The estimated Gaussian mean can be written as
    $\mu_\theta(x_k, k)=\frac{1}{\sqrt{\alpha_k}}\left(x_k-\frac{1-\alpha_k}{\sqrt{1-\bar{\alpha}_k}}\epsilon_\theta (x_k, k)\right)$~,
where $\bar{\alpha}_k=\Pi_{s=1}^k\alpha_s$.

\begin{figure}[t]
   \centering
   \includegraphics[width=0.85\textwidth]{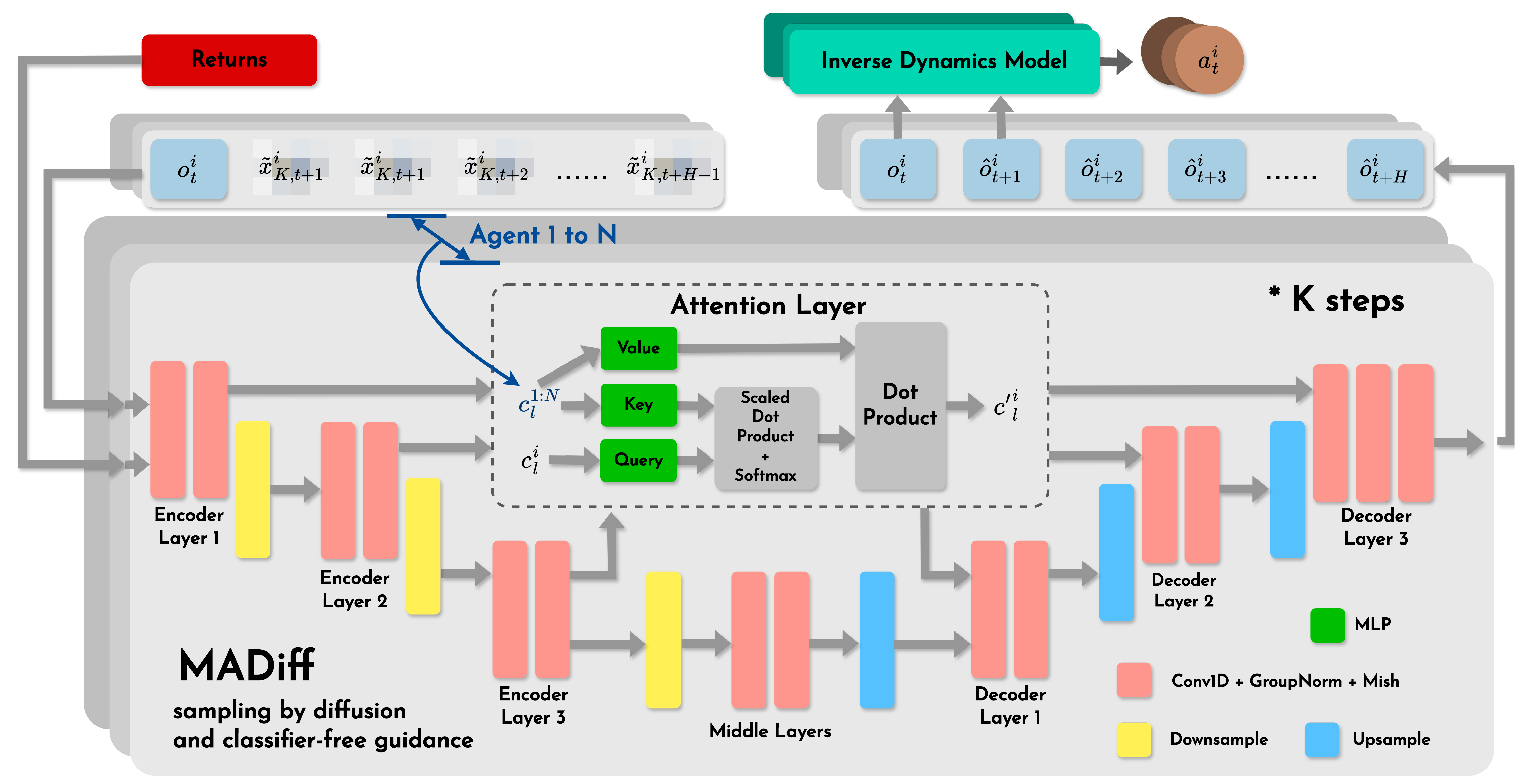}
   \vspace{-4pt}
   \caption{The architecture of \our, which is an attention-based diffusion network framework that performs attention across all agents at every decoder layer of each agent. 
   }
   \label{fig:madiff}
   \vspace{-15pt}
\end{figure}
\subsection{Diffusing Decision Making}
\paragraph{Diffusing over state trajectories and acting with inverse dynamics model.}
Among existing works in single-agent learning, \cite{janner2022planning} chose to diffuse over state-action sequences, so that the generated actions for the current step can be directly used for executing. 
Another choice is diffusing over state trajectories only~\citep{ajay2022conditional}, which is claimed to be easier to model and can obtain better performance due to the less smooth nature of action sequences:
\begin{equation}
    \hat{\tau} := [s_t, \hat{s}_{t+1}, \cdots, \hat{s}_{t+H-1}],
\end{equation}
where $t$ is the sampled time step and $H$ denotes the trajectory length (horizon) modeled by DMs.
But the generated state sequences can not provide actions to be executed during online evaluation.
Therefore, an inverse dynamics model is trained to predict the action $\hat{a}_t$ that makes the state transit from $s_t$ to the generated next state $\hat{s}_{t+1}$:
\begin{equation}\label{eq:inv_dy}
    \hat{a}_t = I_{\phi}(s_t, \hat{s}_{t+1})~.
\end{equation}
Therefore, at every environment step $t$, the agent first plans the state trajectories using an offline-trained DM, and infers the action with the inverse dynamics model.

\paragraph{Classifier-free guided generation.} 
For targeted behavior synthesis, DMs should be able to generate future trajectories by conditioning the diffusion process on an observed state $s_t$ and information $y$.
We use classifier-free guidance~\citep{ho2022classifier} which requires taking $y(\tau)$ as additional inputs for the diffusion model.
Formally, the sampling procedure starts with Gaussian noise 
$\hat{\tau}_K\sim\mathcal{N}(0, \alpha I)$, and diffuse $\hat{\tau}_k$ into $\hat{\tau}_{k-1}$ at each diffusion step $k$. 
Here $\alpha\in[0, 1)$ is the scaling factor used in low-temperature sampling to scale down the variance of initial samples~\citep{ajay2022conditional}.
We use $\tilde{x}_{k,t}$ to denote the denoised state $s_t$ at $k$'s diffusion step.
$\hat{\tau}_k$ denotes the denoised trajectory at $k$'s diffusion step for a single agent:
$\hat{\tau}_k := [s_t, \tilde{x}_{k,t+1}, \cdots, \tilde{x}_{k,t+H-1}]$.
Note that for sampling during evaluations, the first state of the trajectory is always set to the current observed state at all diffusion steps for conditioning, and every diffusion step proceeds with the perturbed noise:
\begin{equation}\label{eq:denoise}
    \hat{\epsilon} := \epsilon_{\theta}(\hat{\tau}_k,\emptyset, k) + \omega(\epsilon_\theta(\hat{\tau}_k, y(\tau), k)-\epsilon_{\theta}(\hat{\tau}_k,\emptyset, k))~,
\end{equation}
where $\omega$ is a scalar for extracting the distinct portions of data with characteristic $y(\tau)$. 
By iterative diffusing the noisy samples, we can obtain a clean state trajectory:
    $\hat{\tau}_0(\tau):=[s_t, \hat{s}_{t+1}, \cdots, \hat{s}_{t+H-1}]$~.

\section{Methodology}
We formulate the problem of MAL as conditional generative modeling:
\begin{equation}
    \max_{\theta}\bbE_{\bm{\tau}\sim \caD}[\log p_{\theta}(\bm{\tau}|\bm{y}(\cdot))]~,
\end{equation}
where $p_{\theta}$ is learned for estimating the conditional data distribution of joint trajectory $\bm{\tau}$, given information $\bm{y}(\cdot)$, such as observations, rewards, and constraints.
When all agents are managed by a centralized controller, \textit{i.e.}, the decisions of all agents are made jointly, we can learn the generative model by conditioning the global information aggregated from all agents $\bm{y}(\bm{\tau})$; otherwise, if we consider each agent $i$ separately and require each agent to make decisions in a decentralized manner, we can only utilize the local information $y^i(\tau^i)$ of each agent $i$, including the private information and the common information shared by all (\textit{e.g.}, team rewards).

\subsection{Multi-Agent Diffusion with Attention}
In order to handle MAL problems, agents must learn to coordinate.
To solve the challenge of modeling the complex inter-agent coordination in the dataset, we propose a novel attention-based diffusion architecture designed to interchange information among agents.

In \fig{fig:madiff}, we illustrate the architecture of \our model.
In detail, we adopt U-Net as the base structure for modeling agents' individual trajectories, which consists of repeated one-dimensional convolutional residual blocks.
The convolution is performed over the time step dimension, and the observation feature dimension is treated as the channel dimension.
To encourage information interchange and improve coordination ability, a critical change is made by adopting attention~\citep{vaswani2017attention} layers before all decoder blocks in the U-Nets of all agents.
Since embedding vectors from different agents are aggregated by the attention operation rather than concatenations, \our is index-free such that the input order of agents can be arbitrary and does not affect the results.

Formally, the input to $l$-th decoder layer in the U-Net of each agent $i$ is composed of two components: the skip-connected feature $c^i_l$ from the symmetric $l$-th encoder layer and the embedding $e^i_l$ from the previous decoder layer.
The computation of attention in \our is conducted on $c^i_l$ rather than $e^i_l$ since in the U-Net structure the encoder layers are supposed to extract informative features from the input data. 
We use ${c'}^i_l$ to denote the skip-connected feature after attention operations which aggregate information across agents.
We adopt the multi-head attention mechanism to fuse the encoded feature ${c'}^i_l$ with other agents' information, which is important in effective multi-agent coordination.

\subsection{Centralized Training Objectives}
Given a multi-agent offline dataset $\caD$,
we train \our which is parameterized through the unified noise model $\epsilon_\theta$ for all agents and the inverse dynamics model $I_\phi^i$ of each agent $i$ with the reverse diffusion loss and the inverse dynamics loss:
\begin{equation}\label{eq:cent-loss}
\begin{aligned}
    \caL(\theta, & \phi) := \sum_{i}\bbE_{(o^i,a^i,o'^i)\in\caD}[\|a^i-I_{\phi}^i(o^i,o'^i)\|^2] \\
    &+ \bbE_{k,\bm{\tau}_0\in\caD,\beta}[\|\epsilon-\epsilon_\theta(\bm{\hat{\tau}}_k, (1-\beta)\bm{y}(\bm{\tau}_0) + \beta\emptyset,k)\|^2]~,
\end{aligned}
\end{equation}
where $\beta$ is sampled from a Bernoulli distribution to balance the training effort on unconditioned and conditioned models.
For training the DM, we sample noise $\epsilon\sim\caN(\bm{0}, \bm{I})$ and a time step $k\sim\caU\{1,\cdots,K\}$, construct a noise corrupted joint state sequence $\bm{\tau}_k$ from $\bm{\tau}$ and predict the noise $\hat{\epsilon}_\theta:=\epsilon_\theta(\bm{\hat{\tau}}_k,\bm{y}(\bm{\tau}_0),k)$. 
Note that the noisy array $\bm{\hat{\tau}}_k$ is applied with the same condition required by the sampling process, as we will discuss in \Secref{sec:cent-decent} in detail.
As for the inverse dynamics training, we sample the observation transitions of each agent to predict the action.

It is worth noting that the choice of whether agents should share their parameters of $\epsilon^i_\theta$ and $I_{\phi^i}$ depends on the homogeneous nature and requirements of tasks. 
If agents choose to share their parameters, only one shared DM and inverse dynamics model are used for generating all agents' trajectories; otherwise, each agent $i$ has extra parameters (\textit{i.e.}, the U-Net and inverse dynamic models) to generate their states and predict their actions.
The attention modules are always shared to incorporate global information into generating each agent's trajectory.

\subsection{Centralized Control or Decentralized Execution}
\label{sec:cent-decent}
\minisection{Centralized control.}
A direct and straightforward way to utilize \our in online decision-making tasks is to have a centralized controller for all agents.
The centralized controller has access to all agents' current local observations and generates all agents' trajectories along with predicting their actions, which are sent to every single agent for acting in the environment.
This is applicable for multi-agent trajectory prediction problems and when interactive agents are permitted to be centralized controlled, such as in team games. 
During the generation process, we sample an initial noise trajectory $\bm{\hat{\tau}}_K$, condition the current joint states of all agents and the global information to utilize $\bm{y}(\bm{\tau}_0)$; following the diffusion step described in \eq{eq:denoise} with $\epsilon_\theta$, we finally sample the joint observation sequence $\bm{\hat{\tau}}_0$ as below:
\begin{equation}\label{eq:diff-ctce}
    \underbrace{[\bm{o}_t, \cdots, \bm{\tilde{x}}_{K,t+H-1}]}_{\bm{\hat{\tau}}_K}
    \xRightarrow{\text{$K$ steps}}
    \underbrace{[\bm{o}_t, \cdots, \bm{\hat{o}}_{t+H-1}]}_{\bm{\hat{\tau}}_0}~,
\end{equation}
where every $\bm{\tilde{x}}_{K, t}\sim\caN(\bm{0},\bm{I})$ is a noise vector sampled from the normal Gaussian. 
After generation, each agent obtains the action through its own inverse dynamics model following \eq{eq:inv_dy} using the current observation $o_t^i$ and the predicted next observation $\hat{o}_{t+1}^i$, and takes a step in the environment.
We highlight that \our provides an efficient way to generate joint actions and the attention module guarantees sufficient feature interactions and information interchange among agents.

\minisection{Decentralized execution with teammate modeling.}
Compared with centralized control, a more popular and widely-adopted setting is that each agent only makes its own decision without any communication with other agents, which is what most current works~\citep{lowe2017multi,rashid2020monotonic,wang2022order} dealt with. In this case, we can only utilize the current local observation of each agent $i$ to plan its own trajectory. To this end, the initial noisy trajectory is conditioned on the current observation of the agent $i$. Similar to the centralized case, by iterative diffusion steps, we finally sample the joint state sequence based on the local observation of agent $i$ as:
\begin{equation}\label{eq:diff-ctde}
     \underbrace{\begin{bmatrix}
     \tilde{x}_{K,t}^0,\cdots,\tilde{x}_{K,t+H-1}^0\\
     \cdots,\\
     o_t^i,\cdots,\tilde{x}_{K,t+H-1}^i\\
     \cdots,\\
     \tilde{x}_{K,t}^N,\cdots,\tilde{x}_{K,t+H-1}^N\\
   \end{bmatrix}}_{\bm{\hat{\tau}}_K^i}
\xRightarrow{\text{$K$ steps}}
   \underbrace{\begin{bmatrix}
     \hat{o}_{t}^0,\cdots,\hat{o}_{t+H-1}^0\\
     \cdots,\\
     o_t^i,\cdots,\hat{o}_{t+H-1}^i\\
     \cdots,\\
     \hat{o}_{t}^N,\cdots,\hat{o}_{t+H-1}^N\\
   \end{bmatrix}}_{\bm{\hat{\tau}}_0^i}~,
\end{equation}
and we can also obtain the action through the agent $i$'s inverse dynamics model as mentioned above.
An important observation is that, the decentralized execution of \our includes teammate modeling such that the agent $i$ infers all others' observation sequences based on its own local observation. We show in experiments that this achieves great performances in various tasks, indicating the effectiveness of teammate modeling and the great ability in coordination. 

\minisection{History-based generation.}
We find DMs are good at modeling the long-term joint distributions, and as a result \our perform better in some cases when we choose to condition on the trajectory of the past history instead of only the current observation. This implies that we replace the joint observation $\bm{o}_t$ in \eq{eq:diff-ctce} as the $C$-length joint history sequence $\bm{h}_t:=[\bm{o}_{t-C},\cdots,\bm{o}_{t-1},\bm{o}_{t}]$, and replace the independent observation $o_t^i$ in \eq{eq:diff-ctde} as the history sequence $h_t^i:=[o_{t-C}^i,\cdots,o_{t-1}^i,o_{t}^i]$ of each agent $i$.
Appendix \se{sec:appendix-illustrate-madiff} illustrates how agents' history and future trajectories are modeled by \our in both centralized control and decentralized execution.

\section{Related Work}

\minisection{Multi-agent Offline RL.}
While offline RL has become an active research topic, only a limited number of works studied offline MARL due to the challenge of offline coordination.
\citet{jiang2021offline} extended BCQ~\citep{fujimoto2019off}, a single-agent offline RL algorithm with policy regularization to multi-agent;
\citet{yang2021believe} developed an implicit constraint approach for offline Q learning, which was found to perform particularly well in MAL tasks;
\citet{pan2022plan} argued the actor update tends to be trapped in local minima when the number of agents increases, and correspondingly proposed an actor regularization method named OMAR. 
All of these Q-learning-based methods naturally have extrapolation error problem~\citep{fujimoto2019off} in offline settings, and their solution cannot get rid of it but only mitigate some. 
As an alternative, MADT~\citep{meng2021offline} formulated offline MARL as return-conditioned supervised learning, and use a similar structure to a previous transformer-based offline RL work~\citep{chen2021decision}.
However, offline MADT learns an independent model for each agent without modeling agent interactions; it relies on the gradient from centralized critics during online fine-tuning to integrate global information into each agent's decentralized policy.
\our not only avoids the problem of extrapolation error, but also achieves the modeling of collaborative information while allowing CTDE in a completely offline training manner.

\minisection{Diffusion Models for Decision-Making.}
There is a recent line of work applying diffusion models (DMs) to decision-making problems such as RL and imitation learning. \citet{janner2022planning} design a diffusion-based trajectory generation model and train a value function to sample high-rewarded trajectories. 
A consequent work~\citep{ajay2022conditional} takes conditions as inputs to the DM, thus bringing more flexibility that generates behaviors that satisfy combinations of diverse conditions.
Another line of work~\citep{wang2022diffusion,hansenestruch2023idql,kang2024efficient} uses the DM as a form of policy, \textit{i.e.}, generating actions conditioned on states, and the training objective behaves as a regularization under the framework of TD-based offline RL algorithms. 
Different from the above, SynthER~\citep{lu2024synthetic} adopts the DM to upsample the rollout data to facilitate learning of any RL algorithms.
All of these existing methods focus on solving single-agent tasks. The proposed \our is structurally similar to \citet{ajay2022conditional}, but includes effective modules to model agent coordination in MAL tasks.

\minisection{Opponent Modeling in MARL.}
Our modeling of teammates can be placed under the larger framework of opponent modeling, which refers to the process by which an agent tries to infer the behaviors or intentions of other agents using its local information. 
There is a rich literature on utilizing opponent modeling in online MARL.
\citet{rabinowitz2018machine} used meta-learning to build three models, and can adapt to new agents after observing their behavior.
SOM~\citep{raileanu2018modeling} uses the agent's own goal-conditioned policy to infer other agents' goals from a maximum likelihood perspective.
LIAM~\citep{papoudakis2021agent} extracts representations of other agents with variational auto-encoders conditioned on the controlled agent's local observations. 
Considering the impact of the ego agent's policy on other agents' policies, LOLA~\citep{foerster2017learning} and following works~\citep{willi2022cola, zhao2022proximal} instead model the parameter update of the opponents.
Different from these methods, \our can use the same generative model to jointly output plans of its own trajectory and predictions of other agents' trajectories and is shown to be effective in offline settings.

\section{Experiments}

In experiments, we are aiming at excavating the ability of \our in modeling the complex interactions among cooperative agents, particularly, whether \our is able to (i) generate high-quality multi-agent trajectories; (ii) appropriately infer teammates' behavior; (iii) learn effective, coordinated policies from offline data. 

\subsection{Task Descriptions}
\label{sec:exp-task}
We conduct experiments on multiple commonly used multi-agent testbeds.
\begin{itemize}[leftmargin=*]
    \item \textbf{Multi-agent particle environments (MPE)}~\citep{lowe2017multi}: multiple 2D particles cooperate to achieve a common goal. 
    \textit{Spread}, three agents start at some random locations and have to cover three landmarks without collisions; \textit{Tag}, three predators try to catch a pre-trained prey opponent that moves faster and needs cooperative containment; \textit{World}, also requires three predators to catch a pre-trained prey, whose goal is to eat the food on the map while not getting caught, and the map has forests that agents can hide and invisible from the outside.
    \begin{itemize} 
        \item \textbf{Datasets}: we use the offline datasets constructed by \citet{pan2022plan}, including four datasets collected by policies of different qualities trained by MATD3~\citep{ackermann2019reducing}, namely, Expert, Medium-Replay (Md-Replay), Medium and Random.
    \end{itemize}
    \item \textbf{Multi-Agent Mujoco (MA Mujoco)}~\citep{peng2021facmac}: independent agents control different subsets of a robot's joints to run forward as fast as possible.
    We use three configurations: \textit{2-agent halfcheetah (2halfcheetah)}, \textit{2-agent ant (2ant)}, and \textit{4-agent ant (4ant)}.
    \begin{itemize}
        \item \textbf{Datasets}: we use the off-the-grid offline dataset~\citep{formanek2023off}, including three datasets with different qualities for each robot control task, \textit{e.g.}, Good, Medium, and Poor.
    \end{itemize}
    \item \textbf{StarCraft Multi-Agent Challenge (SMAC)}~\citep{samvelyan2019starcraft}: a team of either homogeneous or heterogeneous units collaborates to fight against the enemy team that is controlled by the hand-coded built-in StarCraft II AI.
    We cover four maps: \textit{3m}, both teams control three Marines; \textit{2s3z}, both teams control two Stalkers and 3 Zealots; \textit{5m\_vs\_6m (5m6m)}, requires controlling five Marines and the enemy team has six Marines; \textit{8m}, both teams control eight Marines.
        \begin{itemize}
            \item \textbf{Datasets}: we use the off-the-grid offline dataset~\citep{formanek2023off}, including three datasets with different qualities for each map, \textit{e.g.}, Good, Medium, and Poor.
        \end{itemize}
    \item \textbf{Multi-Agent Trajectory Prediction (MATP)}: different from the former offline MARL challenges which should learn the policy for each agent, the MATP problem only requires predicting the future behaviors of all agents, and no decentralized model is needed.
    \begin{itemize}
        \item \textbf{NBA dataset}: the dataset consists of various basketball players' recorded trajectories from 631 games in the 2015-16 season. Following \citet{alcorn2021Baller2Vec++}, we split 569/30/32 training/validation/test games, with downsampling from 25 Hz to 5Hz.
        Different from MARL tasks, other information apart from agents' historical trajectories is available for making predictions, including the ball's historical trajectories, player ids, and a binary variable indicating the side of each player's frontcourt. Each term is encoded and concatenated with diffusion time embeddings as side inputs to each U-Net block.
    \end{itemize}
\end{itemize}

\begin{table}[t]
\caption{The average score on offline MARL tasks. Shaded columns represent our methods. 
The mean and standard error are computed over 5 different seeds.
}
\begin{center}
\resizebox{\linewidth}{!}{
\begin{tabular}{c|c|c|cccccca|a}
    \toprule
    \textbf{Testbed} & \textbf{Task} & \textbf{Dataset} & \textbf{BC} & \textbf{MA-ICQ} & \textbf{MA-TD3+BC} & \textbf{MA-CQL} & \textbf{OMAR} & \textbf{MADT} & \textbf{\our-D} & \textbf{\our-C} \\
    \midrule
    \multirow{12}{*}{MPE} & \multirow{4}{*}{Spread} & Expert & 35.0 $\pm$ 2.6 & 104.0 $\pm$ 3.4 & 108.3 $\pm$ 3.3 & 98.2 $\pm$ 5.2 & \bm{$114.9 \pm 2.6$} & - & 95.0 $\pm$ 5.3 & \blue{\bm{$116.7 \pm 3.0$}} \\
    & & Md-Replay & 10.0 $\pm$ 3.8 & 13.6 $\pm$ 5.7 & 15.4 $\pm$ 5.6 & 20.0 $\pm$ 8.4 & \bm{$37.9 \pm 12.3$} & - & 30.3 $\pm$ 2.5 & \blue{\bm{$42.2 \pm 8.1$}} \\
    & & Medium & 31.6 $\pm$ 4.8 & 29.3 $\pm$ 5.5 & 29.3 $\pm$ 4.8 & 34.1 $\pm$ 7.2 & 47.9 $\pm$ 18.9 & - & \bm{$64.9 \pm 7.7$} & 58.2 $\pm$ 1.7 \\
    & & Random & -0.5 $\pm$ 3.2 & 6.3 $\pm$ 3.5 & 9.8 $\pm$ 4.9 & 24.0 $\pm$ 9.8 & \bm{$34.4 \pm 5.3$} & - & 6.9 $\pm$ 3.1 & 4.3 $\pm$ 2.6 \\
    \cmidrule{2-11}
    & \multirow{4}{*}{Tag} & Expert & 40.0 $\pm$ 9.6 & 113.0 $\pm$ 14.4 & 115.2 $\pm$ 12.5 & 93.9 $\pm$ 14.0 & 116.2 $\pm$ 19.8 & - & \bm{$120.9 \pm 14.6$} & \blue{\bm{$167.6 \pm 18.6$}} \\
    & & Md-Replay & 0.9 $\pm$ 1.4 & 34.5 $\pm$ 27.8 & 28.7 $\pm$ 20.9 & 24.8 $\pm$ 17.3 & 47.1 $\pm$ 15.3 & - & \bm{$62.3 \pm 9.2$} & \blue{\bm{$95.0 \pm 9.7$}} \\
    & & Medium & 22.5 $\pm$ 1.8 & 63.3 $\pm$ 20.0 & 65.1 $\pm$ 29.5 & 61.7 $\pm$ 23.1 & 66.7 $\pm$ 23.2 & - & \bm{$77.2 \pm 10.4$} & \blue{\bm{$132.9 \pm 15.0$}} \\
    & & Random & 1.2 $\pm$ 0.8 & 2.2 $\pm$ 2.6 & 5.7 $\pm$ 3.5 & 5.0 $\pm$ 8.2 & \bm{$11.1 \pm 2.8$} & - & 3.2 $\pm$ 4.0 & 10.7 $\pm$ 4.0 \\
    \cmidrule{2-11}
    & \multirow{4}{*}{World} & Expert & 33.0 $\pm$ 9.9 & 109.5 $\pm$ 22.8 & 110.3 $\pm$ 21.3 & 71.9 $\pm$ 28.1 & 110.4 $\pm$ 25.7 & - & \bm{$122.6 \pm 14.4$} & \blue{\bm{$174.0 \pm 16.8$}} \\
    & & Md-Replay & 2.3 $\pm$ 1.5 & 12.0 $\pm$ 9.1 &  17.4 $\pm$ 8.1 & 29.6 $\pm$ 13.8 & 42.9 $\pm$ 19.5 & - & \bm{$57.1 \pm 10.7$} & \blue{\bm{$83.0 \pm 4.4$}} \\
    & & Medium & 25.3 $\pm$ 2.0 & 71.9 $\pm$ 20.0 & 73.4 $\pm$ 9.3 & 58.6 $\pm$ 11.2 & 74.6 $\pm$ 11.5 & - & \bm{$123.5 \pm 4.5$} & \blue{\bm{$158.2 \pm 6.3$}} \\
    & & Random & -2.4 $\pm$ 0.5 & 1.0 $\pm$ 3.2 & 2.8 $\pm$ 5.5 & 0.6 $\pm$ 2.0 & \bm{$5.9 \pm 5.2$} & - & 2.0 $\pm$ 3.0 & \blue{\bm{$8.1 \pm 3.5$}} \\
    \midrule
    \midrule
    \multirow{9}{*}{\shortstack[c]{MA\\ Mujoco}} & \multirow{3}{*}{2halfcheetah} & Good & 6846 $\pm$ 574 & - & 7025 $\pm$ 439 & - & 1434 $\pm$ 1903 & - & \bm{$8246 \pm 342$} & \blue{\bm{$8514 \pm 336$}} \\
    & & Medium & 1627 $\pm$ 187 & - & \bm{$2561 \pm 82$} & - & 1892 $\pm$ 220 & - & 2207 $\pm$ 23 & $2203 \pm 65$ \\
    & & Poor & 465 $\pm$ 59 & - & 736 $\pm$ 72 & - & 384 $\pm$ 420 & - & \bm{$759 \pm 18$} & \blue{\bm{$760 \pm 15$}} \\
    \cmidrule{2-11}
    & \multirow{3}{*}{2ant} & Good & 2697 $\pm$ 267 & - & 2922 $\pm$ 194 & - & 464 $\pm$ 469 & - & \bm{$2946 \pm 77$} & \blue{\bm{$3069 \pm 60$}} \\
    & & Medium & 1145 $\pm$ 126 & - & 744 $\pm$ 283 & - & 799 $\pm$ 186 & - & \bm{$1211 \pm 69$} & \blue{\bm{$1243 \pm 37$}} \\
    & & Poor & 954 $\pm$ 80 & - & \bm{$1256 \pm 122$} & - & 857 $\pm$ 73 & - & 946 $\pm$ 66 & 1038 $\pm$ 26 \\
    \cmidrule{2-11}
    & \multirow{3}{*}{4ant} & Good & 2802 $\pm$ 133 & - & 2628 $\pm$ 971 & - & 344 $\pm$ 631 & - & \bm{$3080 \pm 38$} & \blue{\bm{$3068 \pm 44$}} \\
    & & Medium & 1617 $\pm$ 153 & - & \bm{$1843 \pm 494$} & - & 929 $\pm$ 349 & - & 1649 $\pm$ 100 & \blue{\bm{$1871 \pm 52$}} \\
    & & Poor & 1033 $\pm$ 122 & - & 1075 $\pm$ 96 & - & 518 $\pm$ 112 & - & \bm{$1295 \pm 57$} & \blue{\bm{$1353 \pm 44$}} \\
    \midrule
    \midrule
    \multirow{12}{*}{SMAC} & \multirow{3}{*}{3m} & Good & 16.0 $\pm$ 1.0 & 18.8 $\pm$ 0.6 & - & \bm{$19.6 \pm 0.3$} & - & 19.1 $\pm$ 0.5 & 19.3 $\pm$ 0.6 & \blue{\bm{$19.9 \pm 0.1$}} \\
    & & Medium & 8.2 $\pm$ 0.8 & 18.1 $\pm$ 0.7 & - & \bm{$18.9 \pm 0.7$} & - & 15.8 $\pm$ 0.4 & 17.3 $\pm$ 0.5 & \blue{\bm{$18.1 \pm 0.6$}} \\
    & & Poor & 4.4 $\pm$ 0.1 & \bm{$14.4 \pm 1.2$} & - & 5.8 $\pm$ 0.4 & - & 4.4 $\pm$ 0.3 & 9.6 $\pm$ 1.7 & 9.5 $\pm$ 0.5 \\
    \cmidrule{2-11}
    & \multirow{3}{*}{2s3z} & Good & 18.2 $\pm$ 0.4 & \bm{$19.6 \pm 0.3$} & - & 19.0 $\pm$ 0.8 & - & 19.3 $\pm$ 0.2 & \bm{$19.6 \pm 0.3$} & \blue{\bm{$19.7 \pm 0.3$}} \\
    & & Medium & 12.3 $\pm$ 0.7 & 17.2 $\pm$ 0.6 & - & 14.3 $\pm$ 2.0 & - & 15.0 $\pm$ 0.6 & \bm{$17.4 \pm 0.2$} & \blue{\bm{$17.6 \pm 0.3$}} \\
    & & Poor & 6.7 $\pm$ 0.3 & \bm{$12.1 \pm 0.4$} & - & 10.1 $\pm$ 0.7 & - & 7.0 $\pm$ 0.3 & 9.8 $\pm$ 0.2 & 10.4 $\pm$ 0.7 \\
    \cmidrule{2-11}
    & \multirow{3}{*}{5m6m} & Good & 16.6 $\pm$ 0.6 & 16.3 $\pm$ 0.9 & - & 13.8 $\pm$ 3.1 & - & 16.7 $\pm$ 0.1 & \bm{$17.8 \pm 0.8$} & \blue{\bm{$18.0 \pm 0.8$}} \\
    & & Medium & 12.4 $\pm$ 0.9 & 15.3 $\pm$ 0.7 & - & 17.0 $\pm$ 1.2 & - & 16.6 $\pm$ 0.2 & \bm{$17.3 \pm 0.5$} & \blue{\bm{$18.0 \pm 0.8$}} \\
    & & Poor & 7.5 $\pm$ 0.2 & 9.4 $\pm$ 0.4 & - & \bm{$10.4 \pm 1.0$} & - & 7.8 $\pm$ 0.4 & 8.9 $\pm$ 0.2 & 10.3 $\pm$ 1.3 \\
    \cmidrule{2-11}
    & \multirow{3}{*}{8m} & Good & 16.7 $\pm$ 0.4 & \bm{$19.6 \pm 0.3$} & - & 11.3 $\pm$ 6.1 & - & 18.4 $\pm$ 0.3 & 19.2 $\pm$ 0.1 & \blue{\bm{$19.8 \pm 0.4$}} \\
    & & Medium & 10.7 $\pm$ 0.5 & 18.6 $\pm$ 0.5 & - & 16.8 $\pm$ 3.1 & - & 18.5 $\pm$ 0.3 & \bm{$18.9 \pm 0.9$} & \blue{\bm{$19.4 \pm 0.9$}} \\
    & & Poor & 5.3 $\pm$ 0.1 & \bm{$10.8 \pm 0.8$} & - & 4.6 $\pm$ 2.4 & - & 4.7 $\pm$ 0.1 & 5.1 $\pm$ 0.1 & 5.1 $\pm$ 0.1 \\
    \bottomrule
\end{tabular}
}
\end{center}
\label{tb:marl-result}
\vspace{-18pt}
\end{table}

\subsection{Compared Baselines and Metrics}
\label{sec:exp-baseline}
For offline MARL experiments, we use the episodic return obtained in online rollout as the performance measure. We include MA-ICQ~\citep{yang2021believe} and MA-CQL~\citep{kumar2020conservative} as baselines on all offline RL tasks.
On MPE, we also include OMAR and MA-TD3+BC~\citep{fujimoto2021minimalist} in baseline algorithms and use the results reported by \citet{pan2022plan}.
On MA Mujoco, baseline results are adopted from \citet{formanek2023off}.
On SMAC, we include MADT~\citep{meng2021offline} as a sequence modeling baseline, while other baseline results are reported by \citet{formanek2023off}.
We implement independent behavior cloning (BC) as a naive supervised learning baseline.

We use distance-based metrics including average displacement error (ADE) $\frac{1}{L\cdot N}\sum_{t=1}^L\sum_{i=1}^{N}\|\hat{o}_t^i-o_t^i\|$ and final displacement error (FDE) $\frac{1}{ N}\sum_{i=1}^{N}\|\hat{o}_L^i-o_L^i\|$, where $L$ is the prediction length~\citep{li2020evolvegraph}.
We also report minADE$_{20}$ and minFDE$_{20}$ as additional metrics to balance the stochasticity in sampling, which are the minimum ADE and FDE among 20 predicted trajectories, respectively. We compare \our with Baller2Vec++~\citep{alcorn2021Baller2Vec++}, an autoregressive MATP algorithm based on the transformer structure and specifically designed for the NBA dataset.

\subsection{Numerical Results}
We reported the numerical results both for the CTDE version of \our (denoted as \our-D) and the centralized version \our (\our-C).
For offline MARL, since baselines are tested in a decentralized style, \textit{i.e.}, all agents independently decide their actions with only local observations, \our-C is not meant to be a fair comparison but to show if \our-D fills the gap for coordination without global information. 
For MATP, due to its centralized prediction nature, \our-C is the only variant involved. 

\minisection{Offline MARL.}
As listed in \tb{tb:marl-result}, \our-D achieves the best result on most of the datasets.
Similar to the single-agent case, direct supervised learning (BC) on the dataset behaves poorly when datasets are mixed quality.
Offline RL algorithms such as MA-CQL that compute conservative values have a relatively large drop in performance when the dataset quality is low.
Part of the reason may come from the fact that those algorithms are more likely to fall into local optima in multi-agent scenarios~\citep{pan2022plan}.
Thanks to the distributional modeling ability of the DM, \our-D generally obtains better or competitive performance compared with OMAR~\citep{pan2022plan} without any design for avoiding bad local optima similar to \citet{pan2022plan}.
On SMAC tasks, \our-D achieves comparable performances, although it is slightly degraded compared with \our-C.

\minisection{MATP on the NBA dataset.}
In \tb{tb:nba}, 
when comparing ADE and FDE, \our-C significantly outperforms the baseline; however, our algorithm only slightly beats baseline for $\text{minADE}_{20}$, and has higher $\text{minFDE}_{20}$. 
We suspect the reason is that Baller2Vec++ has a large prediction variance.
When Baller2Vec++ only predicts one trajectory, a few players’ trajectories deviate from the truth so far that deteriorate the overall ADE and FDE.
When allowing to sample 20 times and calculating the minimum ADE/FDE according to the ground truth, Baller2Vec++ can choose the best trajectory for every single agent, which makes $\text{minADE}_{20}$ and $\text{minFDE}_{20}$ significantly smaller than one-shot metrics.
However, considering it may be not practical to select the best trajectories without access to the ground truth, \our-C is much more stable than Baller2Vec++. 
Predicted trajectories of \our-C and Baller2Vec++ are provided in the Appendix \se{sec:app-vis-nba}.

\begin{minipage}{.45\textwidth}
\captionof{table}{Multi-agent trajectory prediction results on NBA dataset across 3 seeds, given the first step of all agents' positions.}
\vspace{-5pt}
  \begin{center}
  \resizebox{0.95\linewidth}{!}{
    \begin{tabular}{ c|c|ca } 
    \toprule
    \textbf{Traj. Len.} & \textbf{Metric} & \textbf{Baller2Vec++} & \textbf{\our-C} \\
    \midrule
    \multirow{4}{*}{20} & ADE & 15.15 $\pm$ 0.38 & \textbf{7.92 $\pm$ 0.86}\\
    & FDE & 24.91 $\pm$ 0.68 & \textbf{14.06 $\pm$ 1.16}\\
    & minADE$_{20}$ & 5.62 $\pm$ 0.05 & \textbf{5.20 $\pm$ 0.04}\\
    & minFDE$_{20}$ & \textbf{5.60 $\pm$ 0.12} & 7.61 $\pm$ 0.19\\
    \midrule
    \multirow{4}{*}{64} & ADE & 32.07 $\pm$ 1.93 & \textbf{17.24 $\pm$ 0.80}\\
    & FDE & 44.93 $\pm$ 3.02 & \textbf{26.69 $\pm$ 0.13}\\
    & minADE$_{20}$ & 14.72 $\pm$ 0.53 & \textbf{11.40 $\pm$ 0.06}\\
    & minFDE$_{20}$ & \textbf{10.41 $\pm$ 0.36} & 11.26 $\pm$ 0.26\\
    \bottomrule
    \end{tabular}
    }
    \end{center}
    \label{tb:nba}
\end{minipage}
\hfill
\begin{minipage}{.5\textwidth}
   \centering
   \includegraphics[width=\linewidth]{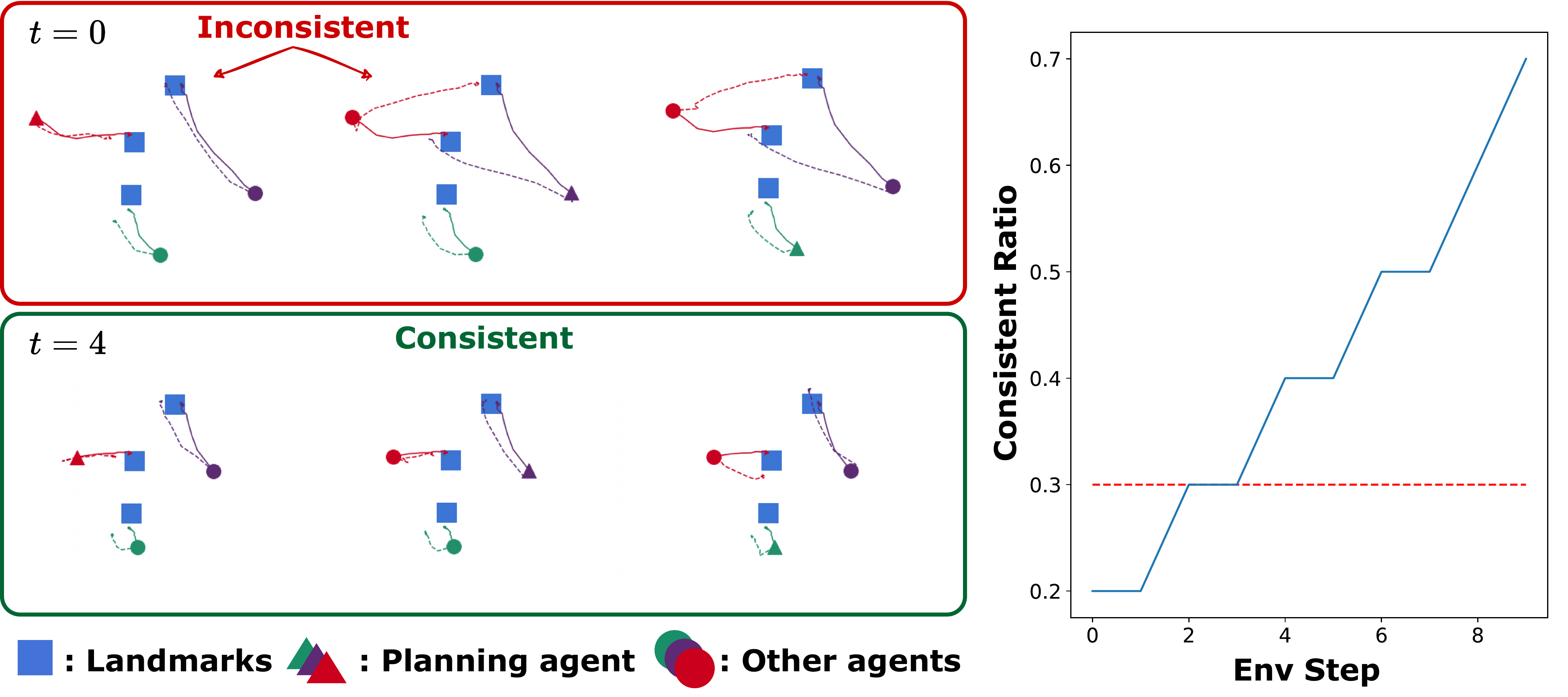}
   \captionof{figure}{Visualization of an episode in the \textit{Spread} task. Solid lines are real rollouts, and dashed lines are DM-planned trajectories.
   }
   \label{fig:mpe-plan-rollout}
\end{minipage}
\vspace{-10pt}

\subsection{Qualitative Analysis on Teammate modeling}
\label{sec:oppo-modeling}
We discuss the quality of teammate modeling as mentioned in \Secref{sec:cent-decent} and how it is related to the decentralized execution scenario.
In \fig{fig:mpe-plan-rollout} left, we visualize an episode generated by \our-D trained on the Expert dataset of \textit{Spread} task. 
The top and bottom rows are snapshots of entities' positions on the initial and intermediate time steps.
The three rows from left to right in each column represent the perspectives of the three agents, red, purple, and green, respectively.
Dashed lines are the planned trajectories for the controlled agent and other agents output by DMs, and solid lines are the real rollout trajectories.
We observe that at the start, the red agent and the purple agent generate \textit{inconsistent} plans, where both agents decide to move towards the middle landmark and assume the other agent is going to the upper landmark. 
At the intermediate time step, when the red agent is close to the middle landmark while far from the uppermost ones, the purple agent altered the planned trajectories of both itself and the red teammate, which makes all agents' plans \textit{consistent} with each other.
This particular case indicates that \our is able to correct the prediction of teammates' behaviors during rollout and modify each agent's own desired goal correspondingly.

In \fig{fig:mpe-plan-rollout} right, we demonstrate that such corrections of teammate modeling are common and can help agents make globally coherent behaviors.
We sample 100 episodes with different initial states and define \textit{Consistent Ratio} at some time step $t$ as the proportion of episodes in which the three agents make consistent planned trajectories. We plot the curve up to step $t=9$, which is approximately halfway through the episode length limit in MPE.
The horizontal red line represents how many portions of the real rollout trajectories are consistent at step $t=9$.
The interesting part is that the increasing curve reaches the red line before $t=9$, and ends up even higher. This indicates that the planned teammates' trajectories are guiding the multi-agent interactions beforehand, which is a strong exemplar of the benefits of \our's teammate modeling abilities.
We also include visualizations of imagined teammate observation sequences in SMAC \textit{3m} task in the Appendix \se{sec:app-vis-smac}.

\subsection{Ablation Study}
Our key argument is that the great coordination ability of \our is brought by the attention modules among individual agents' diffusion networks. We validate this insight through a set of ablation experiments on MPE. We compare \our-D with independent DMs, \textit{i.e.}, each agent learns from corresponding offline data using independent U-Nets without attention. We denote this variant as \our-D-Ind. 
In addition, we also ablate the choice of whether each agent should share parameters of their basic U-Net, noted as Share or NoShare. Without causing ambiguity, we omit the name of \our, and notate the different variants as \textit{D-Share}, \textit{D-NoShare}, \textit{Ind-Share}, \textit{Ind-NoShare}.

\begin{figure}[h!]
\vspace{-8pt}
   \centering
   \includegraphics[width=\textwidth]{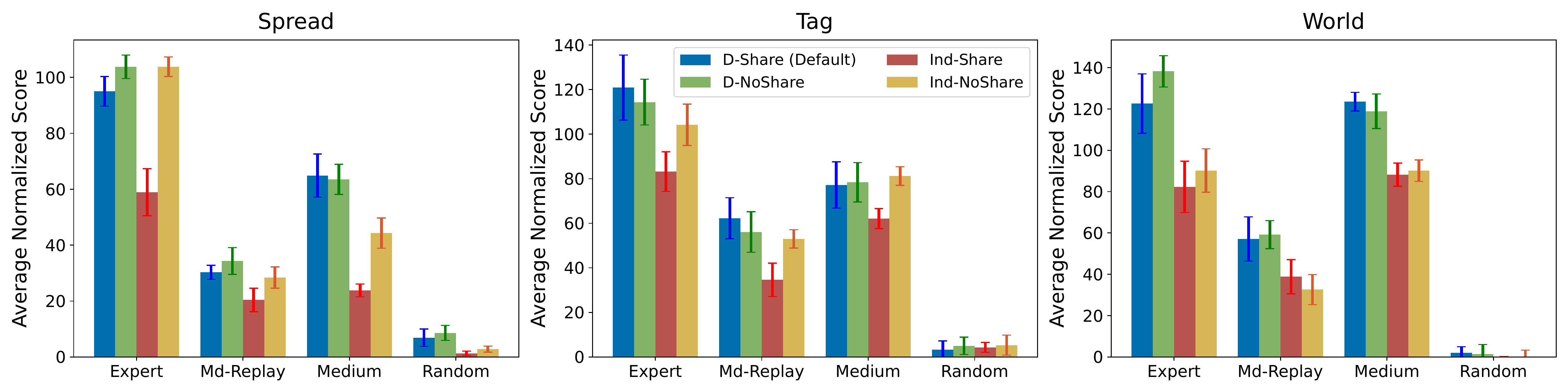}
   \vspace{-18pt}
   \caption{
   The average normalized score of \our ablation variants in MPE tasks.
   The mean and standard error are computed over 5 different seeds.
   }
   \vspace{-10pt}
   \label{fig:ablation-mpe}
\end{figure}

As is obviously observed in \fig{fig:ablation-mpe}, with attention modules, \our-D significantly exceeds that of the independent version on most tasks, justifying the importance of inter-agent attentions. 
The advantage of \our-D is more evident when the task becomes more challenging and the data becomes more confounded, e.g., results on World, where the gap between centralized and independent models is larger, indicating the difficulty of solving offline coordination with independently trained models.
As for the parameter sharing choice, the performance of \our-D-Share and \our-D-NoShare is similar overall.
Since \our-D-Share has fewer parameters, we prefer \our-D-Share, and use it as the default variant to be reported in \tb{tb:marl-result}. 
Another advantage of sharing U-Net parameters is that the trajectories of various agents can be batched together and fed through the network. This not only decreases sampling time but also renders it insensitive to an increasing number of agents. We provide a specific example in Appendix \se{ap-walltime}.

\subsection{Limitations}
\label{sec:limitation}
\minisection{Scalability to many agents.}
\our-D requires each agent to infer all teammates' future trajectories, which is difficult and unnecessary in environments with a large number of agents. Although we have done experiments on a maximum number of 8 agents (SMAC \textit{8m}), \our-D is in general not suitable for scenarios with tens or hundreds of agents. A potential solution is to infer a latent representation of teammates' trajectories.

\minisection{Applicability in highly stochastic environments.}
Several theoretical and empirical studies~\citep{paster2022you,brandfonbrener2022does,chen2021decision} have demonstrated that in offline RL, sequence modeling algorithms tend to underperform Q-learning-based algorithms in environments with high stochasticity.
This is primarily because sequence modeling algorithms are more susceptible to high-reward offline trajectories that are achieved by chance.
Since \our is a sequence modeling algorithm, it shares this weakness.
To assess how much \our is affected by environmental stochasticity, we conducted experiments on the \textit{terran\_5\_vs\_5} map in SMACv2~\citep{ellis2022smacv2}. The design principle of SMACv2 is to add stochasticity to the original SMAC environment, including randomized initial positions and unit types. 
We conducted experiments under four settings: the original version, without position randomness, without unit type randomness, and without both kinds of randomness.
\our performs worse than the Q-learning-based method only when both kinds of stochasticity are present. In all settings, \our outperforms the sequence modeling baseline. Detailed experimental settings and results can be found in Appendix \se{ap-smacv2}.

\section{Conclusion}
\label{sec:conclusion}
In this paper, we propose \our, a novel generative multi-agent learning framework, which is realized with an attention-based diffusion model designed to model the complex coordination among multiple agents. To our knowledge, \our is the first diffusion-based offline multi-agent learning algorithm, which behaves as both a decentralized policy and a centralized controller including teammate modeling, and can be used for multi-agent trajectory prediction. Our experiments indicate strong performance compared with a set of recent offline MARL baselines on a variety of tasks.

\section*{Acknowledgements}
The SJTU team is partially supported by National Key R\&D Program of China (2022ZD0114804), Shanghai Municipal Science and Technology Major Project (2021SHZDZX0102) and National Natural Science Foundation of China (62322603, 62076161).

\bibliographystyle{plainnat}
\bibliography{ref}

\clearpage
\newpage
\appendix
\onecolumn

\section{Outline}
In this appendix, we provide a table to explain the main notations we used in \Secref{ap:notation}. In \Secref{ap:algo}, we give the pseudocode of multi-agent planning and multi-agent trajectory prediction with \our model. In \Secref{sec:appendix-illustrate-madiff}, we demonstrate how multiple agents' trajectories are modeled by \our during centralized control and decentralized execution in an example three-agent environment.
In \Secref{sec:appendix-addinfo-dataset}, we give additional information on offline datasets, including how they are collected, violin plots of return distributions, and a minor issue of MPE dataset.
In \Secref{sec:appendix-baseline-implementation}, we briefly describe the implementation of baseline algorithms and links to related resources.
In \Secref{ap:implementation}, we provide details of the experiments, including the normalization used to compute the average score, the detailed network illustration unrolling each agent's U-Net, crucial hyperparameters, and examples of wall-clock time and resources required for training and sampling from \our. In \Secref{ap:add-results}, we demonstrate and analyze additional experimental results. Specifically, we provide experiment results on SMACv2 to demonstrate how much \our is affected by environmental stochasticity. We also provide ablation results to support the effectiveness of teammate modeling in \our-D, show the quality of teammate modeling by \our-D on SMAC tasks, and visualize predicted multi-player trajectories by \our and the baseline algorithm on the NBA dataset.

\section{Notations}
\label{ap:notation}

\begin{table}[ht]
    \centering
    \caption{List of main notations used in the paper.}
    \vspace{5pt}
    \begin{tabular}{cl}
        \toprule
        \textbf{Notation} & \textbf{Description} \\
        \midrule
        $\mathcal{S}, \mathcal{A}, \Omega$ & state, action, and local observation spaces \\
        $\gamma$ & the discounted factor \\
        $N$ & number of controlled agents \\
        $s_t$ & state at step $t$ \\
        $a^i_t, o^i_t$ & action and local observation of agent $i$ at environment step $t$ \\
        $\bm{a}_t, \bm{o}_t$ & joint action and observation of all agents at environment step $t$ \\
        $r(s, \bm{a})$ & shared reward function \\
        $\bm{\tau}$ & joint trajectory of all agents \\
        $\bm{y}(\bm{\tau})$ & additional conditioning information \\
        $\phi$ & parameters of the inverse dynamics model \\
        $\theta$ & parameters of the diffusion model \\
        $\tilde{x}^i_{k,t}$ & noised observation of agent $i$ at diffusion step $k$ and environment step $t$ \\
        $\bm{\tilde{x}}_{k,t}$ & noised joint observation at diffusion step $k$ and environment step $t$ \\
        $\hat{o}_t^i$ & predicted observation of agent $i$ at environment step $t$ \\
        $\bm{\hat{\tau}}_k$ & noised joint trajectory of all agents at diffusion step $k$ \\
        $h_t^i$ & historical trajectory of agent $i$ up to environment step $t$ \\
        $\bm{h}_t$ & historical joint trajectory of all agents up to environment step $t$ \\

        \bottomrule
    \end{tabular}
    \label{tab:notations}
\end{table}

\section{Algorithm}
\label{ap:algo}
\begin{figure*}[h!]
\centering
\begin{minipage}{\linewidth}
\begin{algorithm}[H]
    \begin{algorithmic}[1]
    \STATE \textbf{Input:} Noise model $\epsilon_{\theta}$, inverse dynamics $I_{\phi}$, guidance scale $\omega$, history length $C$, condition $\boldsymbol{y}$ \\
    \STATE  Initialize $h \leftarrow \texttt{Queue} (\texttt{length}=C)$;\; $t \leftarrow 0$ \hspace{2.25cm} \small{\color{gray}{// Maintain a history of length C}}\\
    \WHILE{not done}
        \STATE Observe joint observation $\bm{o}$;\; $h.\texttt{insert}(\bm{o})$;\; Initialize $\bm{\tau}_K \sim \gN(\bm{0},\alpha \bm{I})$\\
        \FOR{$k = K \ldots 1$}
            \STATE $\bm{\tau}_k[:\texttt{length}(h)] \leftarrow h$ \hspace{1.2cm} \small{\color{gray}{// Constrain plan to be consistent with history}}
            \IF{\texttt{Centralized control}}
                \STATE $\hat{\bm{\epsilon}} \leftarrow \epsilon_\theta(\bm{\tau}_k, k) + \omega (\epsilon_\theta(\bm{\tau}_k, \boldsymbol{y}, k) - \epsilon_\theta(\bm{\tau}_k, k))$ \hspace{1.2cm} \small{\color{gray}{// Classifier-free guidance}}
                \STATE $(\bm{\mu}_{k-1}, \bm{\Sigma}_{k-1}) \leftarrow \texttt{Denoise}(\bm{\tau}_k, \hat{\bm{\epsilon}})$
            \ELSIF{\texttt{Decentralized execution}}
                \FOR{agent $i \in \{1, 2, \dots, N\}$}
                    \STATE $\hat{\epsilon}^i \leftarrow \epsilon_\theta^i(\tau^i_k, k) + \omega (\epsilon_\theta^i(\tau^i_k, y^i, k) - \epsilon_\theta^i(\tau_k^i, k))$ \hspace{1.2cm} \small{\color{gray}{// Classifier-free guidance}}
                    \STATE $(\mu^i_{k-1}, \Sigma^i_{k-1}) \leftarrow \texttt{Denoise}(\tau^i_k, \hat{\epsilon}^i)$
                \ENDFOR
            \ENDIF
            \STATE $\bm{\tau}_{k-1} \sim \gN(\bm{\mu}_{k-1}, \alpha\bm{\Sigma}_{k-1})$ 
        \ENDFOR
        \STATE Extract $(\bm{o}_t, \bm{o}_{t+1})$ from $\bm{\tau}_0$
        \FOR{agent $i \in \{1, 2, \dots, N\}$}
            \STATE $a^i_t \leftarrow f_{\phi^i}(o^i_t, o^i_{t+1})$
        \ENDFOR
        \STATE Execute $\bm{a}_t$ in the environment;\; $t \leftarrow t + 1$
    \ENDWHILE
    \end{algorithmic}
    \caption{Multi-Agent Planning with \our}
    \label{alg:ma-plan}
\end{algorithm}
\end{minipage}
\end{figure*}

\begin{figure*}[h!]
\centering
\begin{minipage}{\linewidth}
\begin{algorithm}[H]
    \begin{algorithmic}[1]
    \STATE \textbf{Input:} Noise model $\epsilon_{\theta}$, guidance scale $\omega$, condition $\boldsymbol{y}$, historical joint observations $h$ with length $C$, predict horizon $H$ \\
    \STATE Initialize $\bm{\tau}_K \sim \gN(\bm{0},\alpha \bm{I})$ \\
    \FOR{$k = K \ldots 1$}
        \STATE $\bm{\tau}_k[:C] \leftarrow h$ \hspace{1.2cm} \small{\color{gray}{// Constrain prediction to be consistent with history}}
        \STATE $\hat{\bm{\epsilon}} \leftarrow \epsilon_\theta(\bm{\tau}_k, k) + \omega (\epsilon_\theta(\bm{\tau}_k,\boldsymbol{y}, k) - \epsilon_\theta(\bm{\tau}_k, k))$ \hspace{1.2cm} \small{\color{gray}{// Classifier-free guidance}}
        \STATE $(\bm{\mu}_{k-1}, \bm{\Sigma}_{k-1}) \leftarrow \texttt{Denoise}(\bm{\tau}_k, \hat{\bm{\epsilon}})$
        \STATE $\bm{\tau}_{k-1} \sim \gN(\bm{\mu}_{k-1}, \alpha\bm{\Sigma}_{k-1})$ 
    \ENDFOR
    \STATE Extract prediction $(\bm{o}_C, \bm{o}_{C+1}, \dots, \bm{o}_{C+H-1})$ from $\bm{\tau}_0$
    \end{algorithmic}
    \caption{Multi-Agent Trajectory Prediction with \our}
    \label{alg:ma-pred}
\end{algorithm}
\end{minipage}
\end{figure*}

\section{Illustration of Multi-agent Trajectory Modeling}
\label{sec:appendix-illustrate-madiff}
\begin{figure*}[h]
  \centering
  \begin{subfigure}[b]{0.36\linewidth}
    \includegraphics[width=\textwidth]{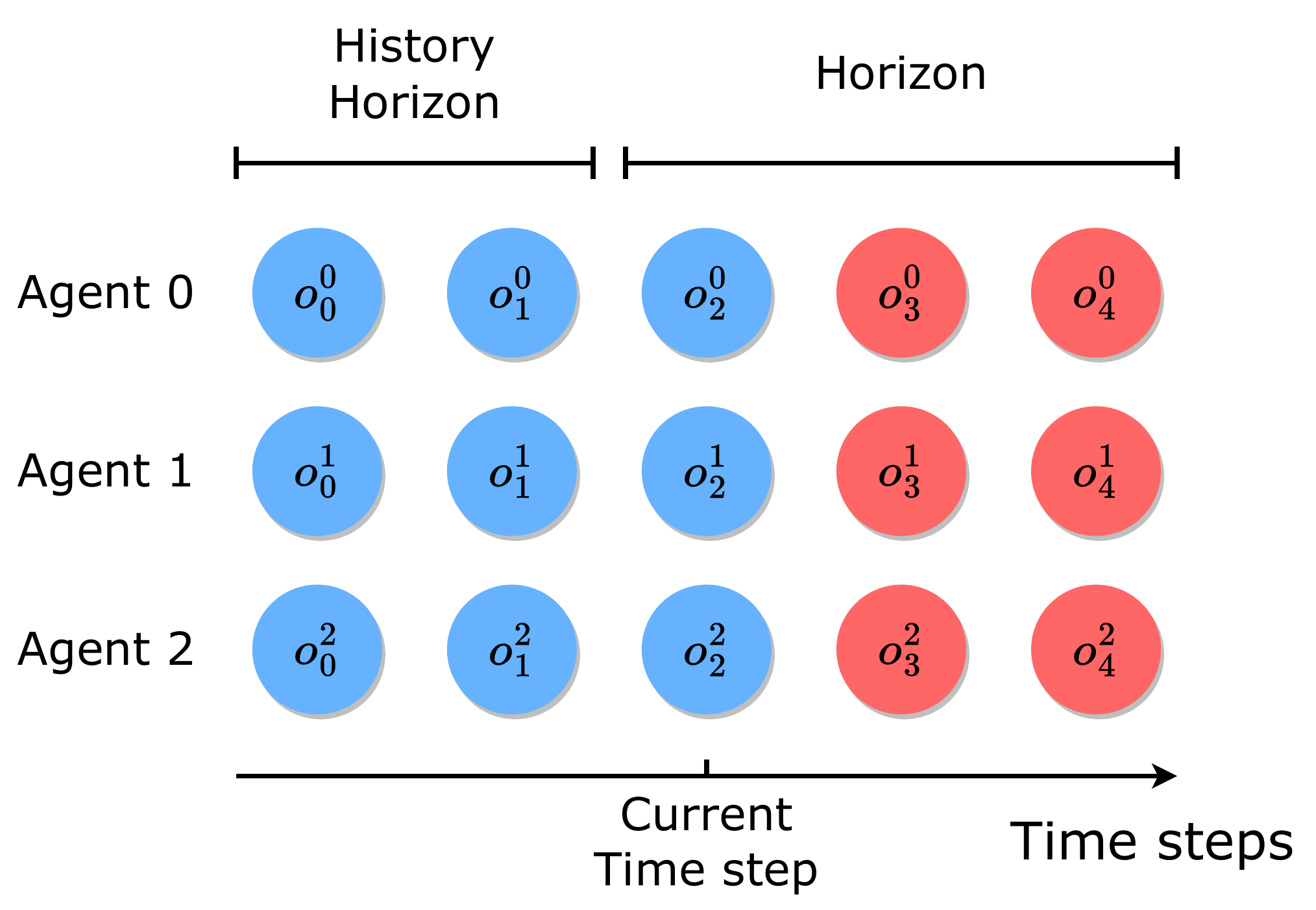}
    \caption{\our in centralized control.}
    \label{fig:appendix-cond-madiff-c}
  \end{subfigure}
  \hfill
  \begin{subfigure}[b]{0.55\linewidth}
    \includegraphics[width=\textwidth]{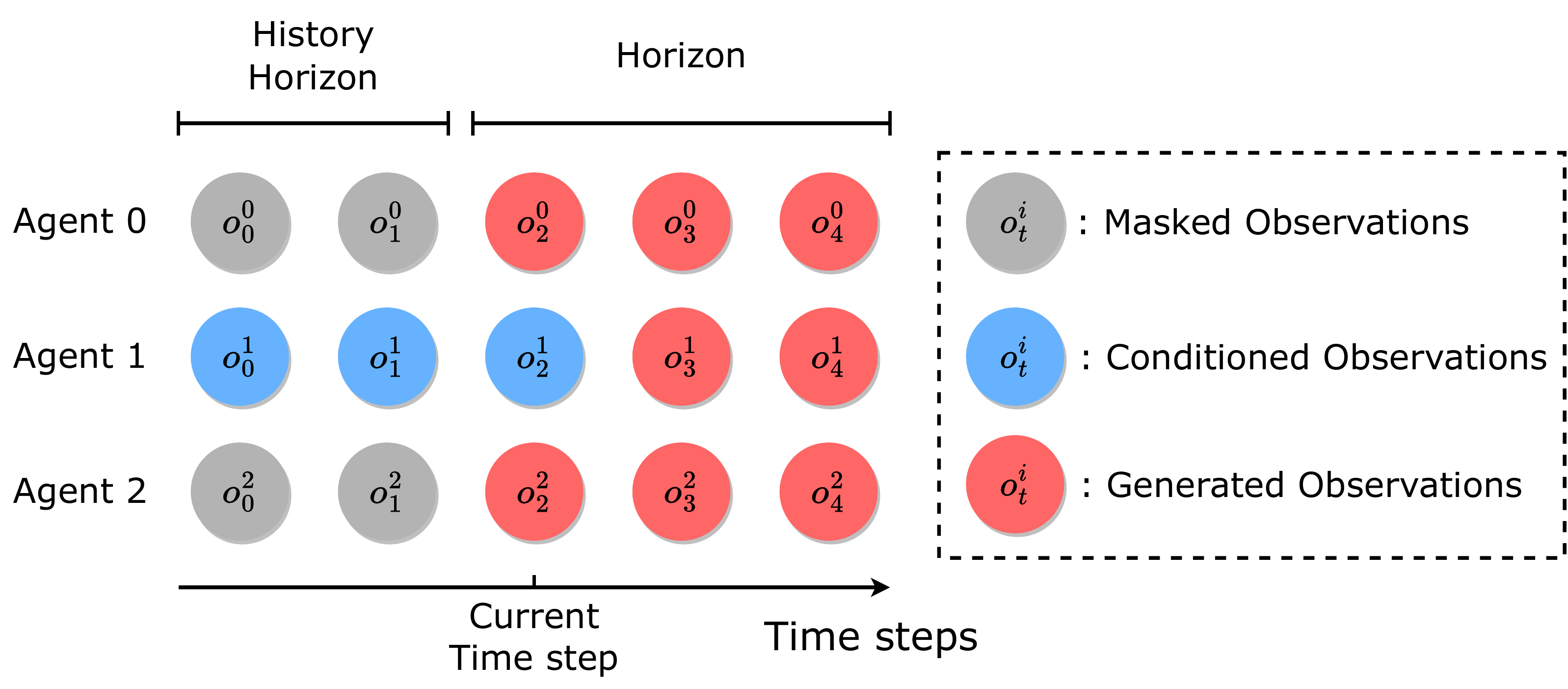}
    \caption{\our in decentralized execution.}
    \label{fig:appendix-cond-madiff-d}
  \end{subfigure}
  \caption{Illustration of how agents' observations are modelled by \our in a three-agent environment. Note that figure (b) shows the situation when Agent 1 is taking action during decentralized execution.}
  \label{fig:appendix-cond-madiff}
\end{figure*}

To provide a better understanding of how multiple agents' observations are modeled by \our in centralized control and decentralized execution scenarios, we show illustrative examples in a typical three-agent environment in \fig{fig:appendix-cond-madiff}. 
If the environment allows for centralized control, we can condition \our on all agents' historical and current observations, and let the model sample all agents' future trajectories as a single sample, as shown in \fig{fig:appendix-cond-madiff-c}. Then the current and next observations are sent to the inverse dynamics model for action prediction. If only decentralized execution is permitted, as shown in \fig{fig:appendix-cond-madiff-d}, agent 1 can only condition the model on its own information. The historical and current observations of other agents are masked when performing conditioning. \our now not only generates agent 1's own future trajectories but also predicts the current and future observations of the other two agents. Due to the joint modeling of all agents during training, such predictions are also reasonable and can be considered as a form of teammate modeling from agent 1's perspective. Although teammate modeling is not directly used in generating agent 1's ego actions, it can help agent 1 refine its planned trajectories to be consistent with the predictions of others.

\section{Additional Information on offline datasets}
\label{sec:appendix-addinfo-dataset}
\subsection{MPE Datasets}
For MPE experiments, we use datasets and a fork of environment\footnote{\url{https://github.com/ling-pan/OMAR}} provided by OMAR~\citep{pan2022plan}. They seem to be using an earlier version of MPE where agents can receive different rewards. For example, in the \textit{Spread} task, team reward is defined using the distance of each landmark to its closest agent, which is the same for all agents. But when an agent collides with others, it will receive the team reward minus a penalty term. The collision reward has been brought into the team reward in the official repository since this commit\footnote{\url{https://github.com/openai/multiagent-particle-envs/commit/6ed7cac026f0eb345d4c20232bafa1dc951c68e7}}. However, the fork provided by OMAR still uses a legacy version. For fair and proper comparisons, we use OMAR's dataset and environment where all baseline models are trained and evaluated. 

We have to note that different rewards for agents only happen at very few steps, which might not contradict the fully cooperative setting much. For example, OMAR's expert split of the \textit{Spread} dataset consists of 1M steps, and different rewards are recorded only at less than 1.5\% (14929) steps. 

\subsection{MA Mujoco Datasets}
For MA Mujoco experiments, we adopt the off-the-grid dataset~\cite{formanek2023off} and use \textit{Good}, \textit{Medium} and \textit{Poor} datasets for each task.
Each dataset is collected by three independently trained MA-TD3 policies, and a small amount of exploration noise is added to the policies for enhanced behavioral diversity.

\begin{figure}[h]
  \centering
  \begin{subfigure}[b]{0.48\linewidth}
    \includegraphics[width=\textwidth]{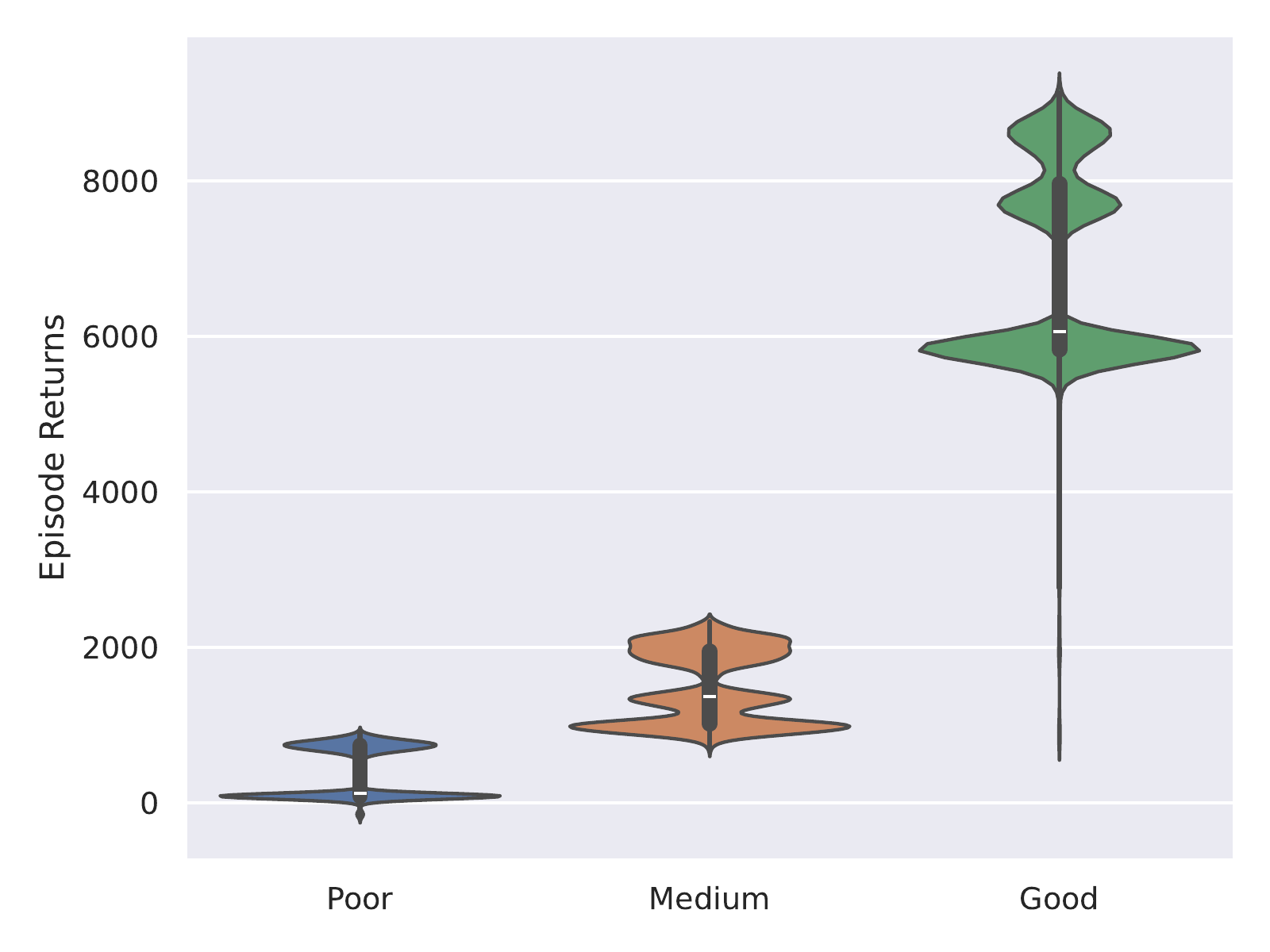}
    \caption{MA Mujoco 2halfcheetah.}
    \label{fig:appendix-violin-2halfcheetah}
  \end{subfigure}
  \hfill
  \begin{subfigure}[b]{0.48\linewidth}
    \includegraphics[width=\textwidth]{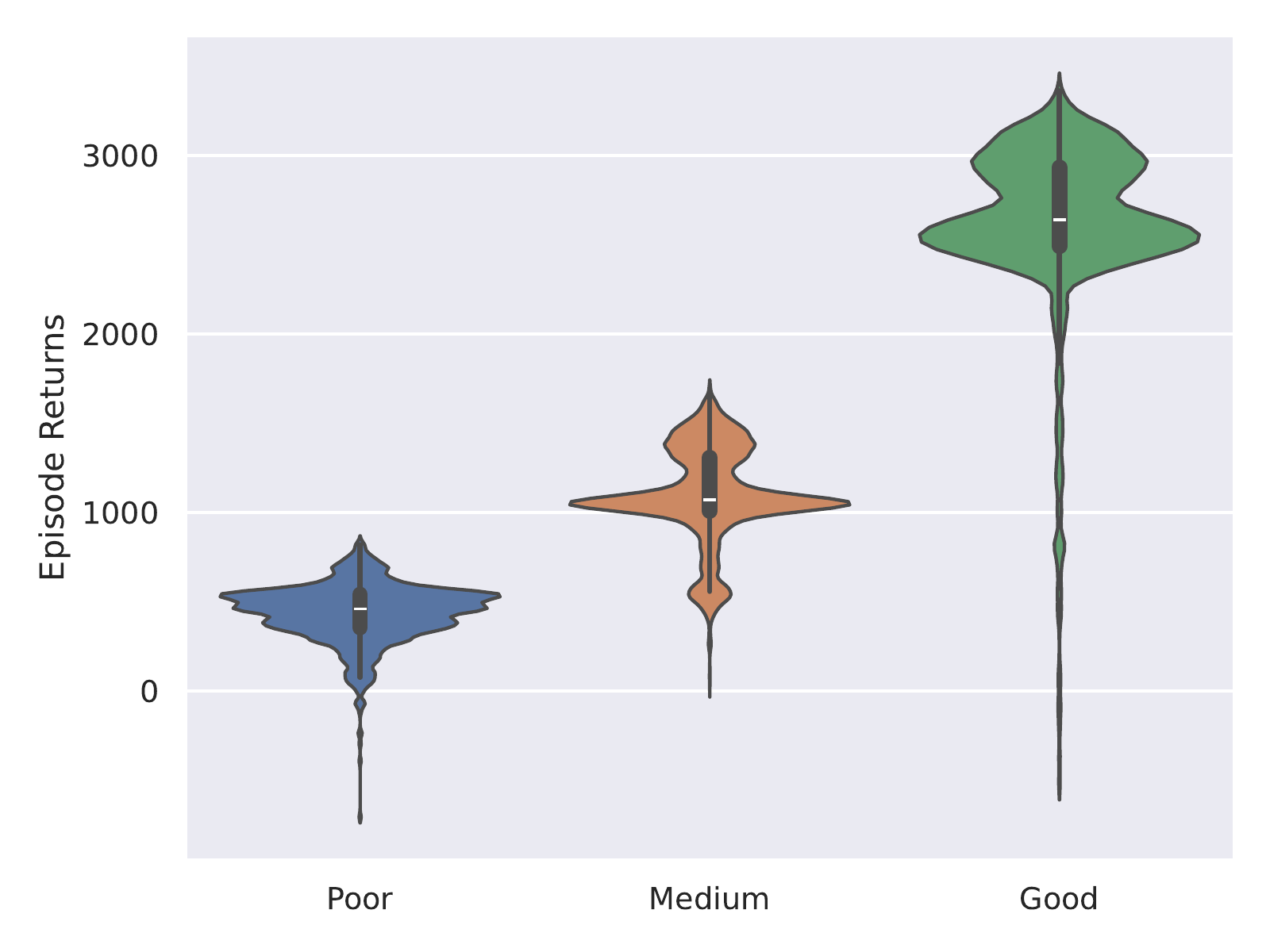}
    \caption{MA Mujoco 2ant.}
    \label{fig:appendix-violin-2ant}
  \end{subfigure}
  \begin{subfigure}[b]{0.48\linewidth}
    \includegraphics[width=\textwidth]{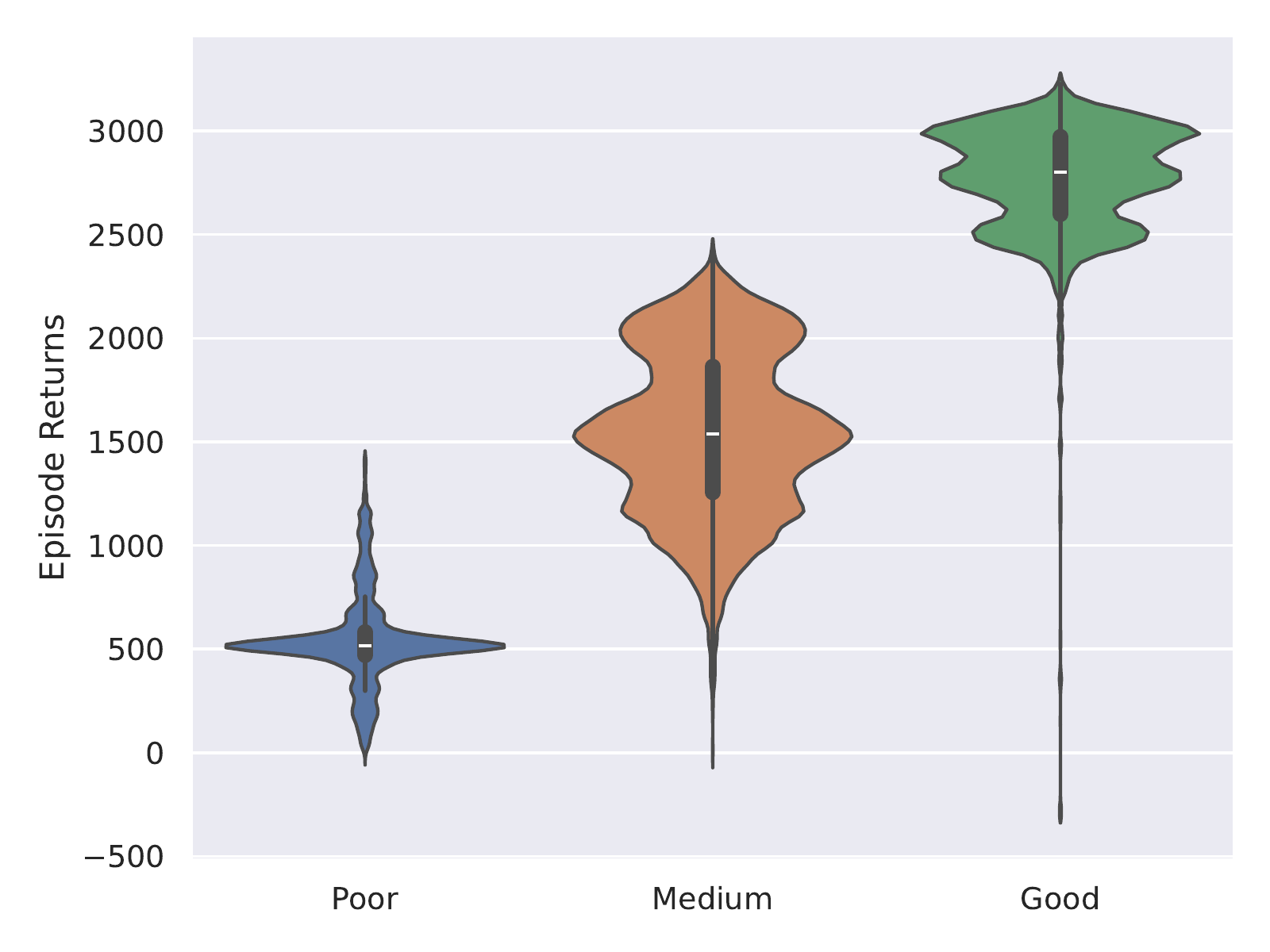}
    \caption{MA Mujoco 4ant.}
    \label{fig:appendix-violin-4ant}
  \end{subfigure}
\caption{Violin plots of returns in MA Mujoco datasets.}
\label{fig:appendix-violin-mamujoco}
\end{figure}

For visualizations of the distribution of episode returns in each dataset, we provide violin plots of all datasets we used in \fig{fig:appendix-violin-mamujoco}.

\begin{figure}[h]
  \centering
  \begin{subfigure}[b]{0.48\linewidth}
    \includegraphics[width=\textwidth]{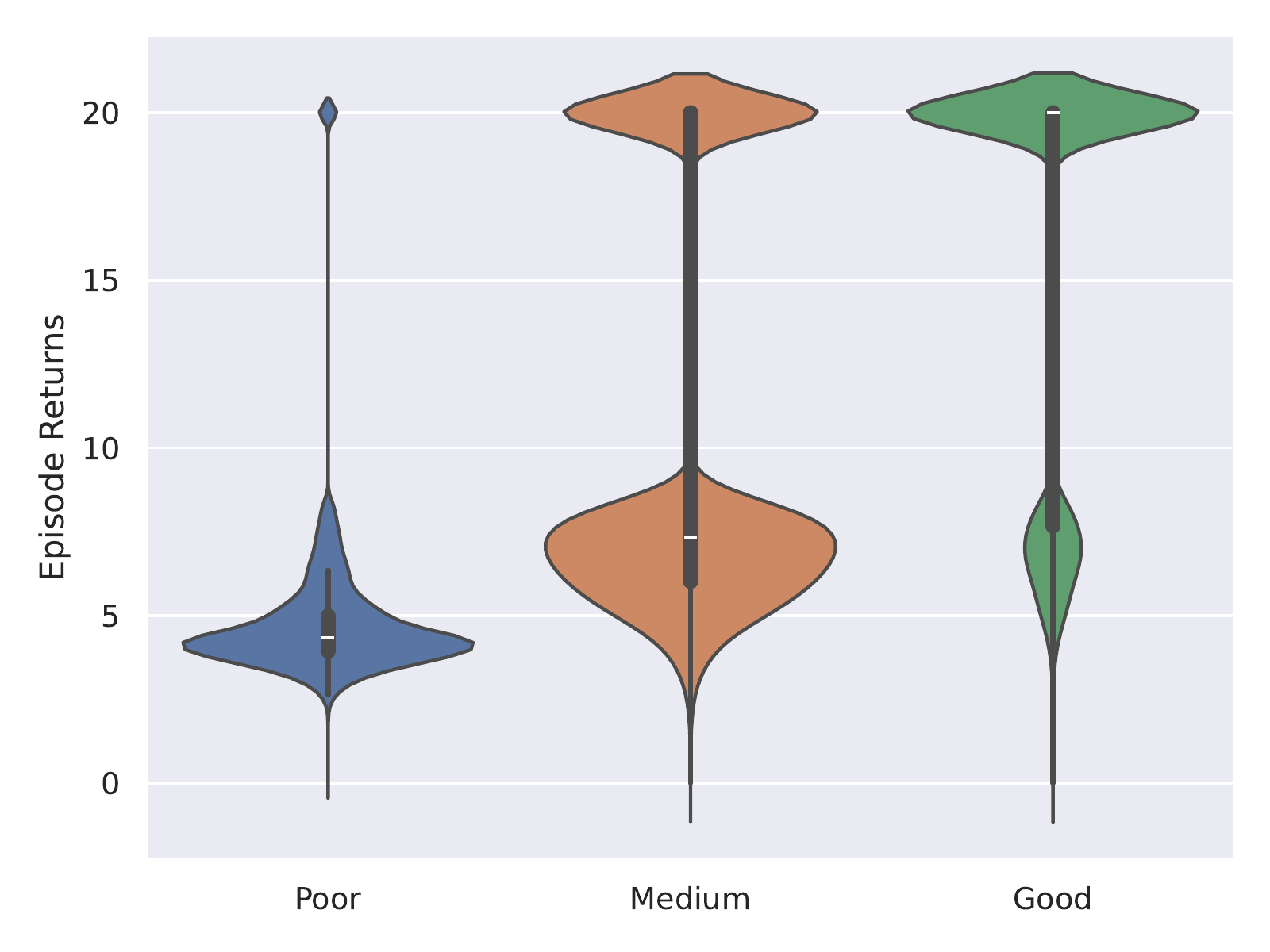}
    \caption{SMAC 3m.}
    \label{fig:appendix-violin-3m}
  \end{subfigure}
  \hfill
  \begin{subfigure}[b]{0.48\linewidth}
    \includegraphics[width=\textwidth]{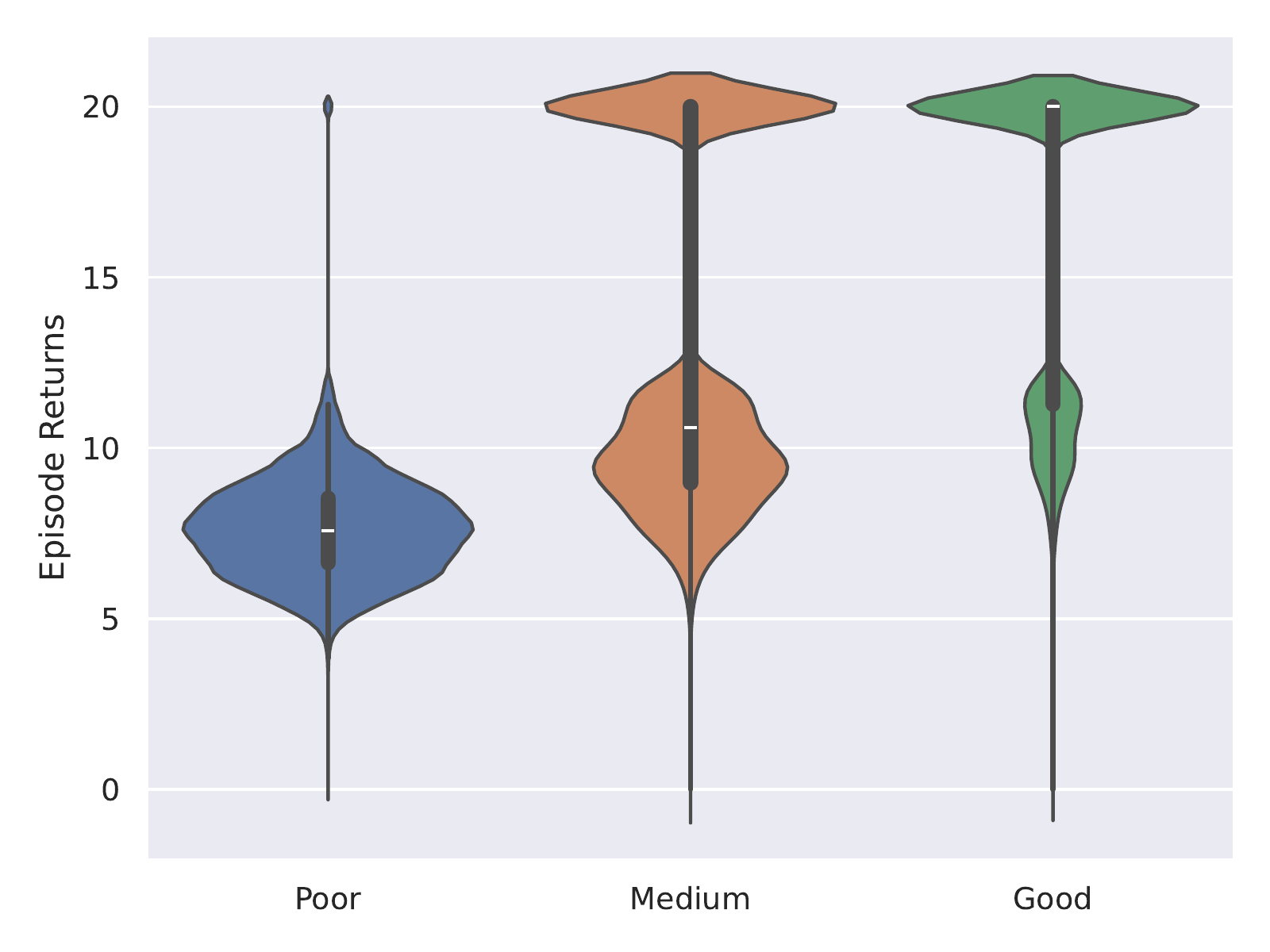}
    \caption{SMAC 5m\_vs\_6m.}
    \label{fig:appendix-violin-5m6m}
  \end{subfigure}
  \begin{subfigure}[b]{0.48\linewidth}
    \includegraphics[width=\textwidth]{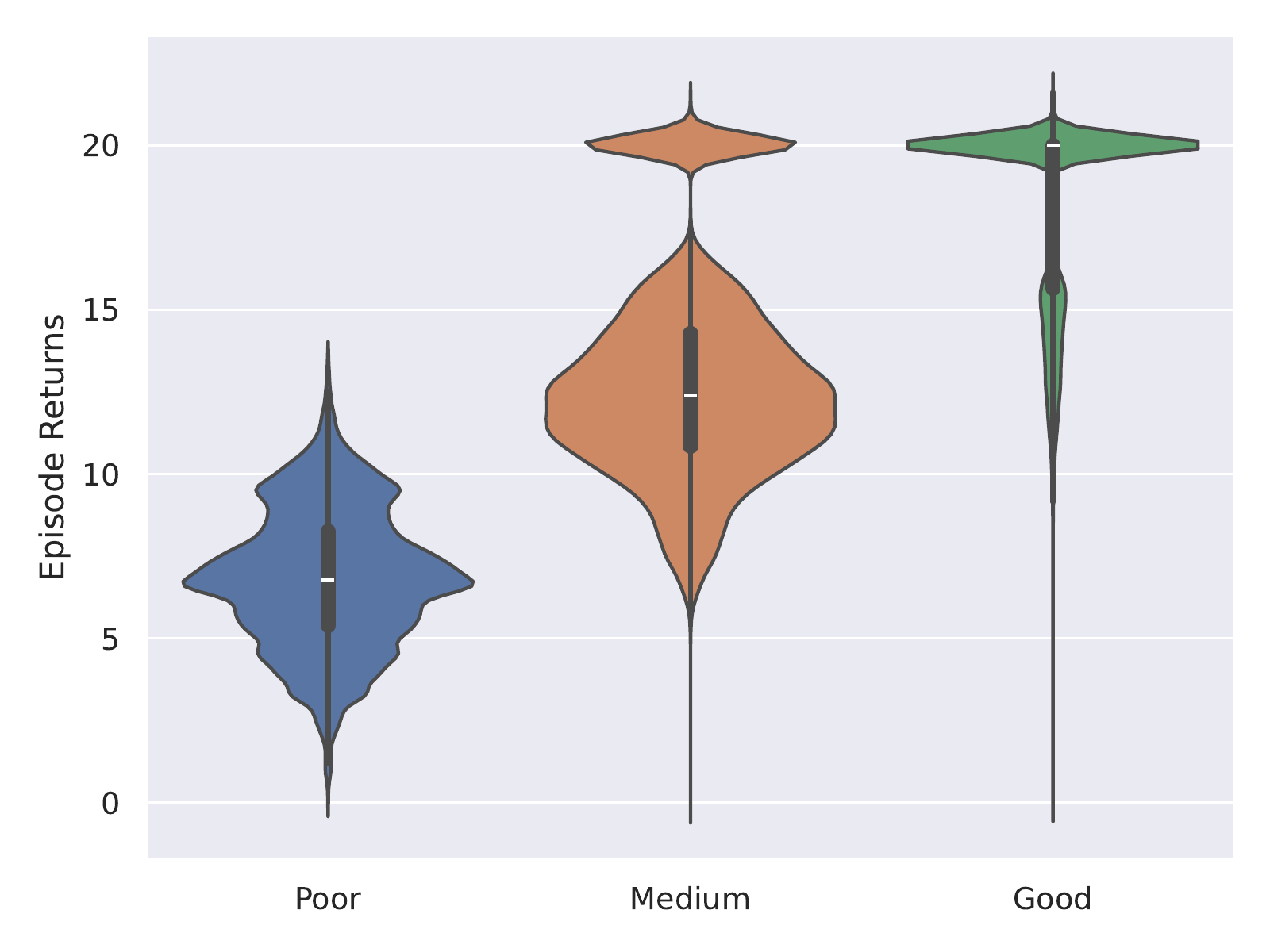}
    \caption{SMAC 2s3z.}
    \label{fig:appendix-violin-2s3z}
  \end{subfigure}
  \hfill
  \begin{subfigure}[b]{0.48\linewidth}
    \includegraphics[width=\textwidth]{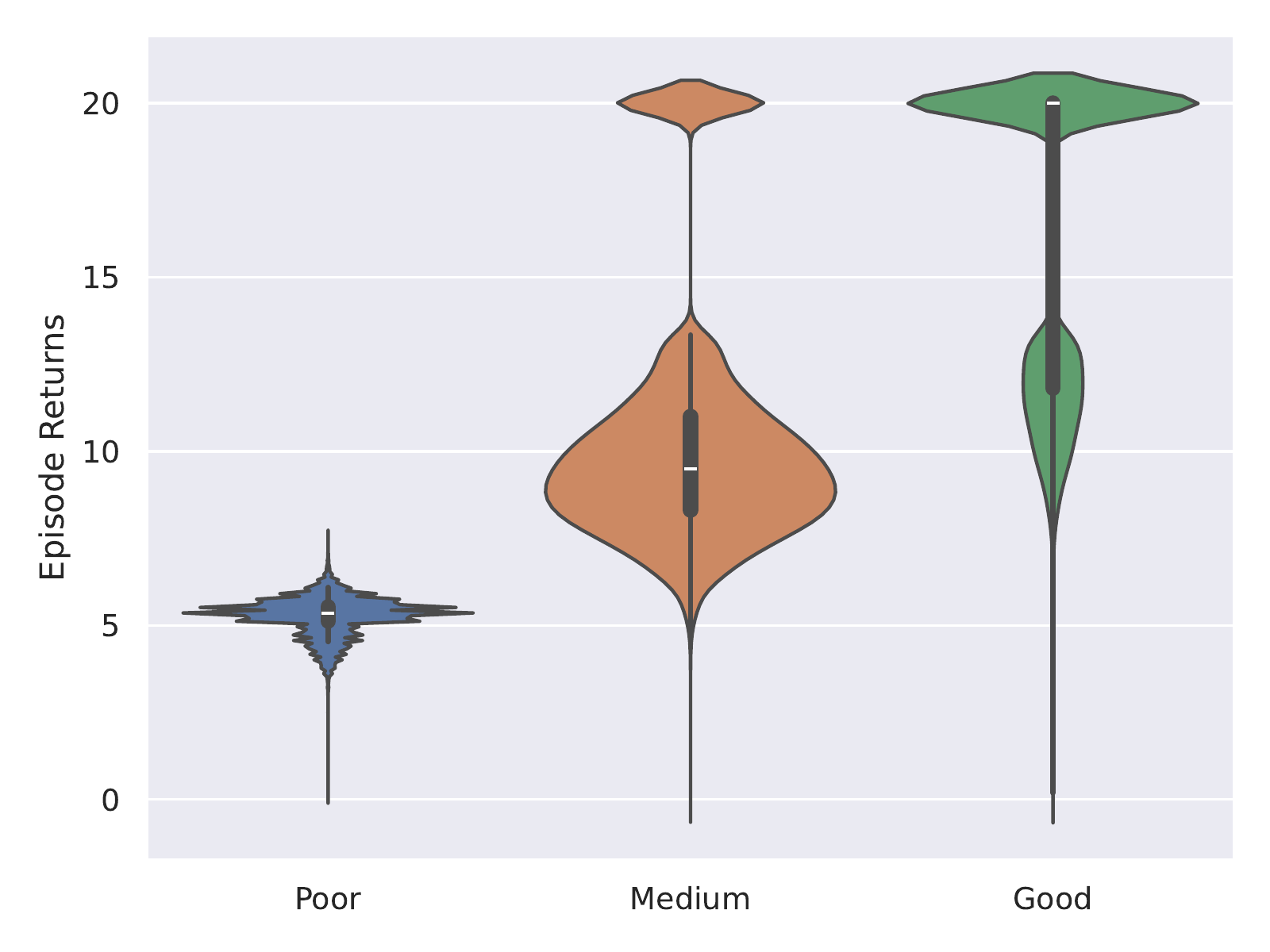}
    \caption{SMAC 8m.}
    \label{fig:appendix-violin-8m}
  \end{subfigure}
\caption{Violin plots of returns in SMAC datasets.}
\label{fig:appendix-violin-smac}
\end{figure}

\subsection{SMAC Datasets}
For SMAC experiments, we adopt the off-the-grid dataset~\citep{formanek2023off} and use \textit{Good}, \textit{Medium} and \textit{Poor} datasets for each map. 
Each dataset is collected by three independently trained QMIX policies, and a small amount of exploration noise is added to the policies for enhanced behavioral diversity.

For visualizations of the distribution of episode returns in each dataset, we provide violin plots of all datasets we used in \fig{fig:appendix-violin-smac}.

\section{Baseline Implementations}
\label{sec:appendix-baseline-implementation}
Here we briefly describe how the baseline algorithms are implemented. 
For MATP experiments, we use the implementation from the official repository of Baller2Vec++\footnote{\url{https://github.com/airalcorn2/baller2vecplusplus}}. Baseline results on MPE datasets are borrowed from \citet{pan2022plan}. According to their paper, they build all algorithms upon a modified version of MADDPG\footnote{\url{https://github.com/shariqiqbal2810/maddpg-pytorch}}, which uses decentralized critics for all methods. Baselines on SMAC datasets are implemented by \citet{formanek2023off}, and the performances are adopted from their reported benchmark results. The open-sourced implementation and hyperparameter settings can be found in the official repository\footnote{\url{https://github.com/instadeepai/og-marl}}.

\section{Implementation Details}
\label{ap:implementation}
\subsection{Score Normalization}
The average scores of MPE tasks in \tb{tb:marl-result} are normalized by the expert and random scores on each task. Denote the original episodic return as $S$, then the normalized score $S_{\text{norm}}$ is computed as
\begin{equation*}
    S_{\text{norm}} = 100 \times (S - S_{\text{random}})/(S_{\text{expert}} - S_{\text{random}})~,
\end{equation*}
which follows \citet{pan2022plan} and \citet{fu2020d4rl}. The expert and random scores on Spread, Tag, and World are \{516.8, 159.8\}, \{185.6, -4.1\}, and \{79.5, -6.8\}, respectively.

\subsection{Detailed Network Architecture}
\begin{figure*}[h!]
    \centering
    \includegraphics[width=0.8\linewidth]{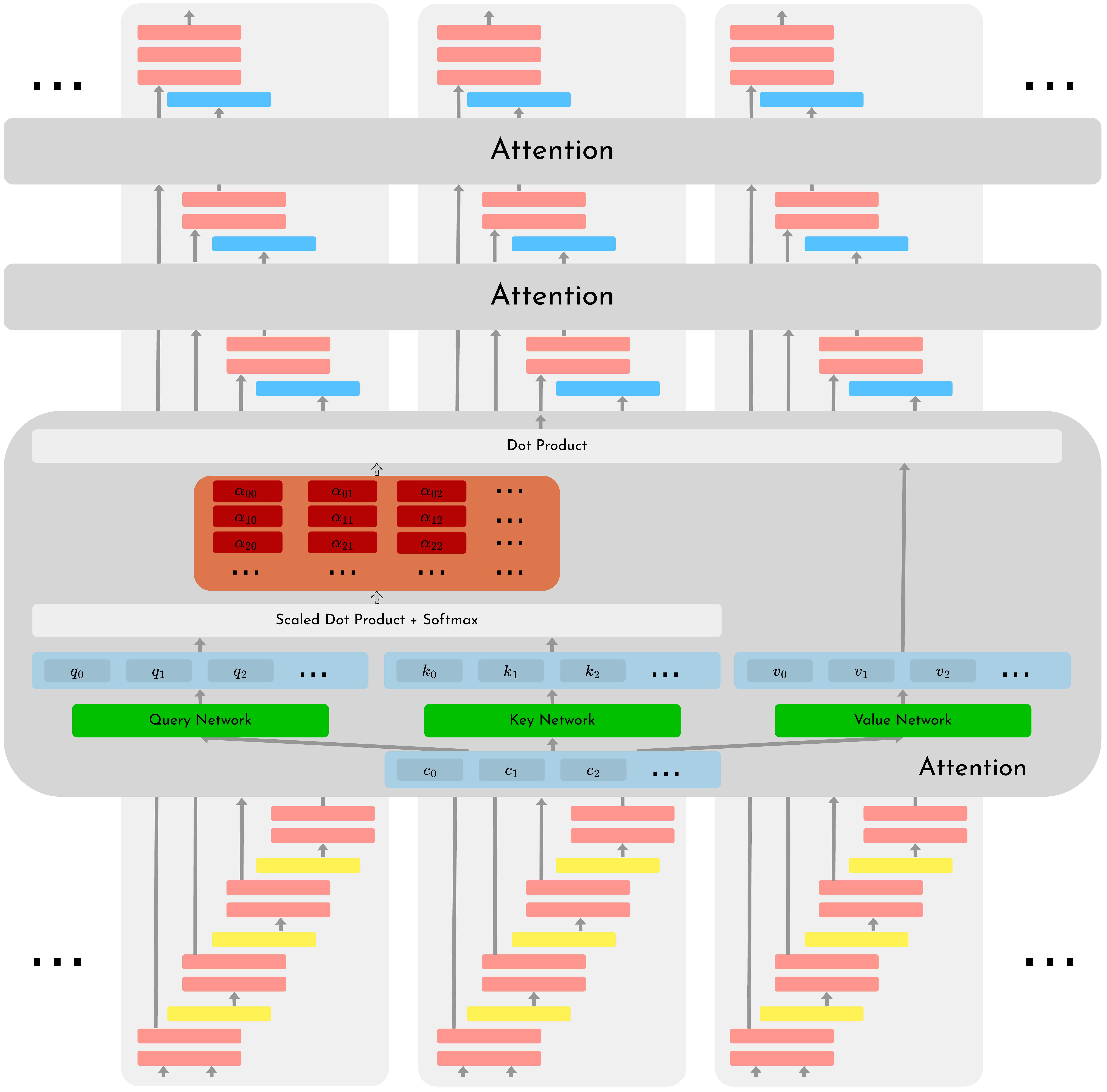}
    \caption{The detailed architecture of \our. Each agent's U-Net is unrolled and lined up in the horizontal direction.}
    \label{fig:madiff-appendix}
\end{figure*}

In \fig{fig:madiff-appendix}, we unroll the U-Net structure of different agents.

We describe the computation steps of attention among agents in formal. Each agent's local embedding $c^i$ is passed through the key, value, and query network to form $q^i$, $k^i$, and $v^i$, respectively. Then the dot product with scaling is performed between all agents' $q^i$ and $k^i$, which is followed by a Softmax operation to obtain the attention weight $\alpha^{ij}$.
Each $\alpha^{ij}$ can be viewed as the importance of $j$-th agent to the $i$-th agent at the current time step. The second dot product is carried out between the weight matrix and the value embedding $v^i$ to get $\hat{c}^i$ after multi-agent feature interactions. Then $\hat{c}^i$ is skip-connected to the corresponding decoder block.
The step-by-step computation of multi-agent attention in \our can be written as
\begin{equation*}
    \begin{aligned}
        q^i=f_{\text{query}}(c^i),\,k^i&=f_{\text{key}}(c^i),\,v^i=f_{\text{value}}(c^i)~; \\
        \alpha^{ij}&=\frac{\exp(q^i k^j / \sqrt{d_k})}{\sum_{p=1}^N\exp(q^i k^p / \sqrt{d_k})}~; \\
        \hat{c}^i&=\sum_{j=1}^N \alpha^{ij}v^j~,
    \end{aligned}
\end{equation*}
where $d_k$ is the dimension of $k^i$.

\subsection{Hyperparameters}
\label{sec:ap-hyperparameter}
We list the key hyperparameters of \our we used in \tb{tb:hypers-madiff-mpe}, \tb{tb:hypers-madiff-mamujoco}, and \tb{tb:hypers-madiff-smac}. In all of our experiments, we use a scaling factor of 0.5 and $\beta$ of 0.25. Return scale is the normalization factor used to divide the conditioned return before input to the diffusion model. The rough range of the return scale can be determined by the return distributions of the training dataset. 
We only tune the guidance weight $\omega$, return scale, planning horizon $H$, and history horizon. We tried the guidance weight of \{1.0, 1.2, 1.4, 1.6\}, and found that different choices do not significantly affect final performances, we chose 1.2 for all experiments. 
For MPE tasks, we find it unnecessary to condition on history observation sequence; thus, we set all history horizons to zero.
\begin{table*}[h!]
\caption{Hyperparameters of \our on MPE datasets.}
\label{tb:hypers-madiff-mpe}
\vspace{5pt}
  \centering
  \resizebox{\textwidth}{!}{
    \begin{tabular}{l|c|c|c|c|c|c|c|c|c|c|c|c}
    \hline
    TestBed & \multicolumn{4}{c|}{Spread} & \multicolumn{4}{c|}{Tag} & \multicolumn{4}{c}{World} \\
    \hline
    Dataset & Expert & Md-Replay & Medium & Random & Expert & Md-Replay & Medium & Random & Expert & Md-Replay & Medium & Random \\
    \hline
    Return scale & 350 & \multicolumn{2}{c|}{200} & 50 & 350 & \multicolumn{2}{c|}{200} & 50 & 200 & \multicolumn{2}{c|}{100} & 10 \\
    \hline
    Learning rate & \multicolumn{12}{c}{2e-4} \\
    \hline
    Guidance scale $\omega$ & \multicolumn{12}{c}{1.2} \\
    \hline
    Planning horizon $H$ & \multicolumn{12}{c}{24} \\
    \hline
    History horizon & \multicolumn{12}{c}{0} \\
    \hline
    Batch size & \multicolumn{12}{c}{32} \\
    \hline
    Diffusion steps $K$ & \multicolumn{12}{c}{200} \\
    \hline
    Reward discount $\gamma$ & \multicolumn{12}{c}{0.99} \\
    \hline
    Optimizer & \multicolumn{12}{c}{Adam Optimizer} \\
    \hline
    \end{tabular}
  }
\end{table*}

\begin{table}[h!]
\caption{Hyperparameters of \our on MA Mujoco datasets.}
\label{tb:hypers-madiff-mamujoco}
\vspace{5pt}
  \centering
  \resizebox{0.9\textwidth}{!}{
    \begin{tabular}{l|c|c|c|c|c|c|c|c|c}
    \hline
    TestBed & \multicolumn{3}{c|}{2halfcheetah} & \multicolumn{3}{c|}{4ant} & \multicolumn{3}{c}{2ant} \\
    \hline
    Dataset & Good & Medium & Poor & Good & Medium & Poor & Good & Medium & Poor \\
    \hline
    Return scale & 1000 & 300 & 100 & 380 & 320 & 150 & 380 & 320 & 150 \\
    \hline
    Learning rate & \multicolumn{9}{c}{2e-4} \\
    \hline
    Guidance scale $\omega$ & \multicolumn{9}{c}{1.2} \\
    \hline
    Planning horizon $H$ & \multicolumn{9}{c}{10} \\
    \hline
    History horizon & \multicolumn{9}{c}{18} \\
    \hline
    Batch size & \multicolumn{9}{c}{32} \\
    \hline
    Diffusion steps $K$ & \multicolumn{9}{c}{200}\\
    \hline
    Reward discount $\gamma$ & \multicolumn{9}{c}{0.99} \\
    \hline
    Optimizer & \multicolumn{9}{c}{Adam Optimizer} \\
    \hline
    \end{tabular}
  }
\end{table}

\begin{table}[h!]
\caption{Hyperparameters of \our on SMAC datasets.}
\label{tb:hypers-madiff-smac}
\vspace{5pt}
  \centering
  \resizebox{\textwidth}{!}{
    \begin{tabular}{l|c|c|c|c|c|c|c|c|c|c|c|c}
    \hline
    TestBed & \multicolumn{3}{c|}{3m} & \multicolumn{3}{c|}{2s3z} & \multicolumn{3}{c|}{5m6m} & \multicolumn{3}{c}{8m} \\
    \hline
    Dataset & Good & Medium & Poor & Good & Medium & Poor & Good & Medium & Poor & Good & Medium & Poor \\
    \hline
    Return scale & \multicolumn{2}{c|}{20} & 8 & \multicolumn{2}{c|}{20} & 12 & \multicolumn{2}{c|}{20} & 10 & \multicolumn{2}{c|}{20} & 8 \\
    \hline
    Learning rate & \multicolumn{12}{c}{2e-4} \\
    \hline
    Guidance scale $\omega$ & \multicolumn{12}{c}{1.2} \\
    \hline
    Planning horizon $H$ & \multicolumn{12}{c}{4} \\
    \hline
    History horizon & \multicolumn{12}{c}{20} \\
    \hline
    Batch size & \multicolumn{12}{c}{32} \\
    \hline
    Diffusion steps $K$ & \multicolumn{12}{c}{200}\\
    \hline
    Reward discount $\gamma$ & \multicolumn{12}{c}{1.0} \\
    \hline
    Optimizer & \multicolumn{12}{c}{Adam Optimizer} \\
    \hline
    \end{tabular}
  }
\end{table}

\subsection{Computing Resources and Wall Time}
\label{ap-walltime}
\begin{figure}[h]
    \centering
    \includegraphics[width=0.5\linewidth]{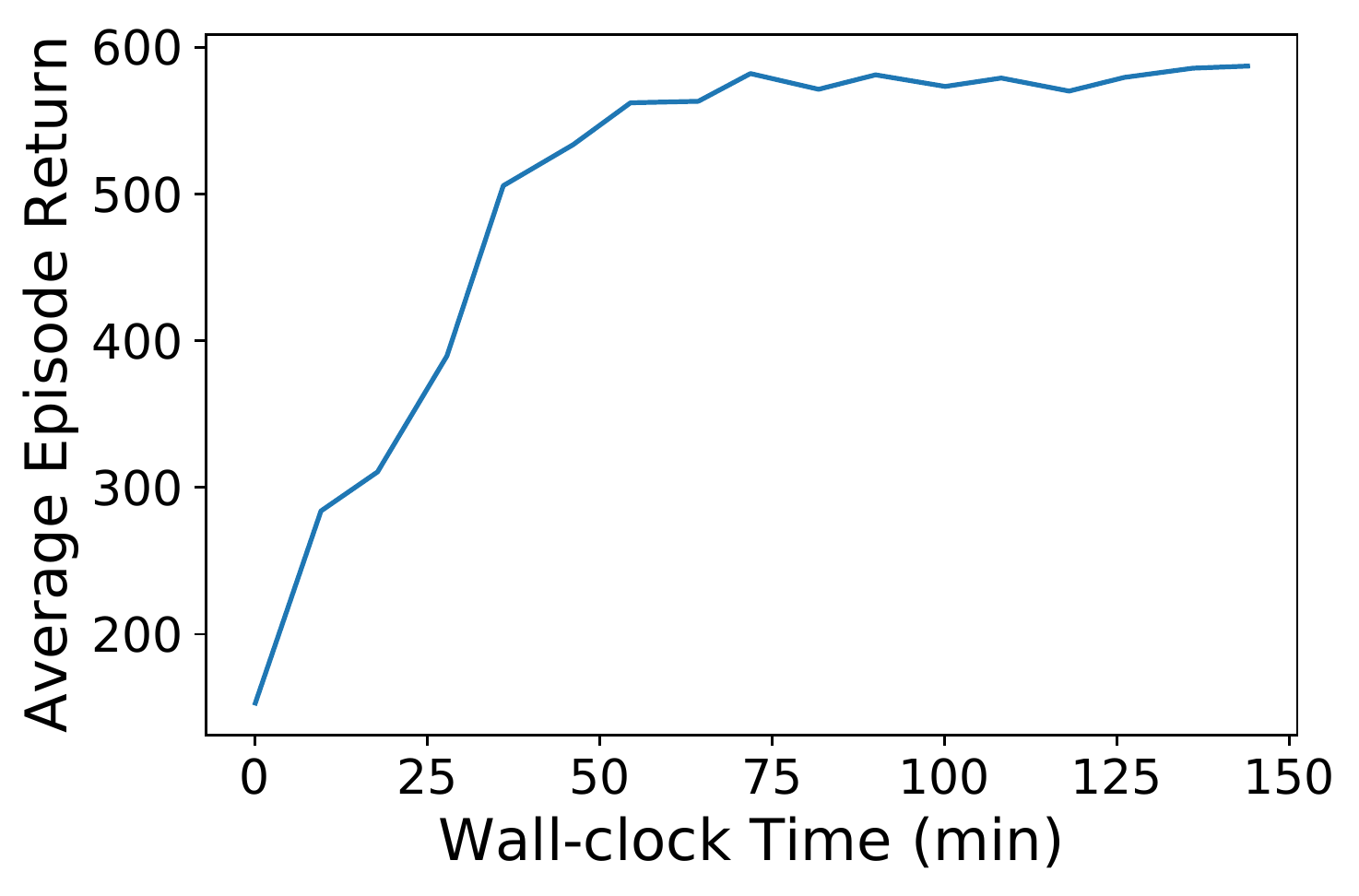}
    \caption{Wall-clock time and corresponding average episode return (average over 10 episodes) during training \our-C for MPE Spread task.}
    \label{fig:mpe-walltime}
\end{figure}

The training of \our does not involve an iterative process, and thus, the training time is not related to the total number of diffusion steps. Thanks to the property that the sum of two independent Gaussian random variables remains a Gaussian, the multistep forward process can be written in a closed form~\citep{ho2020denoising}:
\begin{equation}
    q(\bm{\hat{\tau}}_k|\bm{\tau}_0)=\mathcal{N}(\bm{\hat{\tau}}_k;\sqrt{\bar{\alpha}_t}\bm{\tau}_0, (1-\bar{\alpha}_t)\bm{I})~.
\end{equation}
Therefore, the $k$-th step noisy trajectory in \eq{eq:cent-loss} can be easily sampled from the Gaussian distribution above without an iterative process. 
We provide a concrete example to illustrate the time and resources required for training \our. On a server equipped with an AMD Ryzen 9 5900X (12 cores) CPU and an RTX 3090 GPU, we trained the \our-C model on the Expert dataset from the MPE Spread task, achieving convergence in approximately one hour. The curve depicting Wall-clock time spent on training and the corresponding model performance is shown in \fig{fig:mpe-walltime}.

\begin{table}[h]
\caption{The wall-clock time spent when generating multi-agent trajectories. We fix the dimension of observation space to 88 and use DDIM of 15 steps during sampling. The history horizon is set to 20, and the planning horizon is 8. The results are obtained on a server with an AMD Ryzen 9 5900X (12 cores) CPU and an RTX 3090 GPU, and are averaged over 1000 trials. The computation time does not increase much with the number of agents thanks to GPU-accelerated computing.}
\vspace{5pt}
\label{tb:ab-sampletime}
\centering
\resizebox{0.7\textwidth}{!}{
\begin{tabular}{ccccc}
\toprule
\textbf{Num. Agents (Incl. Ego)} & 8 & 16 & 32 & 64 \\
\midrule
Wall clock Time & 124.25 ms & 126.90 ms & 127.65 ms & 127.35 ms \\
\bottomrule
\end{tabular}
}
\end{table}
In \tb{tb:ab-sampletime}, we showcase the time required for sampling multi-agent trajectories with \our as the number of agents increases.
We can see that the sampling time does not differ much when generating different number of trajectories.
Since we use shared U-Net models for all agents in our experiments, different agents' trajectories can be batched together and passed through the network. Therefore, using GPU-accelerated computing, the second part does not cost much more time than predicting each agent's trajectory during inference.

\section{Additional Experimental Results}
\label{ap:add-results}
\subsection{SMACv2 Experiments}
\label{ap-smacv2}
To understand how much \our is affected by environmental stochasticity, we conducted experiments on the \textit{terran\_5\_vs\_5} map in SMACv2~\citep{ellis2022smacv2}.
SMACv2 is built upon SMAC with a focus on higher stochasticity. Specifically, in SMACv2, the unit types and agent start positions are randomized at the beginning of each episode. As each agent can only observe a nearby area, such randomness results in increased stochasticity in environment transitions. There are two different types of starting positions, reflect and surround. In reflect settings, the map is splitted into two sides. Allied units and enemy units are randomly and uniformly spawned on different sides.
In surround settings, allied units are spawned at the center of the map, and enemy units are randomly stationed along the four diagonals. In \textit{terran\_5\_vs\_5}, there are three different unit types: marine, marauder, and medivac. The default sampling probabilities of these three types are 0.45, 0.45 and 0.1.

We design four settings with different degree of stochasicity: the original version, without position randomness (w/o PR), without unit type randomness (w/o TR), and without both kinds of randomness (w/o PR\&TR). To reduce position randomness, we only use surrounding settings. Note that this does not mean the staring positions of all units are fixed, since enemy units are still randomized along the four diagonals. To remove unit type randomness, we set all units to be marines. The dataset for the original version is the \textit{terran\_5\_vs\_5} Replay dataset from \citet{formanek2023off}. Datasets for other three stochasicity settings were collected by ourselves. 
We partially trained three MAPPO\footnote{\url{https://github.com/marlbenchmark/on-policy}} models in each setting.
Each model was then used to collect 500 episodes, resulting in a dataset comprising 1500 episodes for each setting.

Three algorithms are benchmarked under these four settings: MAICQ, which represents the state-of-the-art in Q-learning-based algorithms; MADT, a representative multi-agent sequence modeling baseline; and \our-D. Results are presented in \tb{tb:ap-smacv2}.
We can see that \our-D performs worse than MAICQ only when both kinds of stochasticity are present.
As the environmental randomness diminishes, \our-D's performance gradually catches up with and surpasses MAICQ.
In all settings, \our-D outperforms MADT.
\begin{table}[h]
\caption{The average score on different settings of SMACv2 \textit{terran\_5\_vs\_5}. Shaded columns represent our method. The mean and standard error are computed over 3 different seeds.}
\label{tb:ap-smacv2}
\begin{center}
\begin{tabular}{ c|c c a } 
 \toprule
 \textbf{Setting} & \textbf{MAICQ} & \textbf{MADT} & \textbf{\our-D} \\
 \midrule
    Original & \bm{$13.7 \pm 1.7$} & 8.2 $\pm$ 0.2 & 10.1 $\pm$ 0.8 \\
    w/o PR & \bm{$16.0 \pm 1.6$} & 14.3 $\pm$ 0.8 & \bm{$16.1 \pm 0.3$} \\
    w/o TR & \bm{$18.4 \pm 0.5$} & 14.6 $\pm$ 0.3 & \bm{$18.6 \pm 0.2$}  \\
    w/o PR\&TR & 17.3 $\pm$ 0.3 & 16.8 $\pm$ 0.3 & \bm{$18.5 \pm 0.2$} \\
 \bottomrule
\end{tabular}
\end{center}
\end{table}

\subsection{Effectiveness of Teammate Modeling}
To investigate whether teammate modeling can lead to performance improvements during decentralized execution, we conduct ablation experiments on MPE Spread datasets. We compare \our-D with its variant that adopts the same network architecture but masks the diffusion loss on other agents' trajectories during training. We denote the variant as \our-D w/o TM. The results are presented in \tb{tb:ab-oppo}, which show that teammate modeling results in notable performance improvements on all four levels of datasets.
\begin{table}[htbp]
\caption{Ablation results of teammate modeling on MPE Spread datasets across 3 seeds. }
\label{tb:ab-oppo}
\begin{center}
\begin{tabular}{ c|c a } 
 \toprule
 \textbf{Dataset} & \textbf{\our-D w/o TM} & \textbf{\our-D} \\
 \midrule
    Expert & 93.4 $\pm$ 3.6 & \textbf{98.4 $\pm$ 12.7} \\
    Medium & 35.4 $\pm$ 6.6 & \textbf{53.2 $\pm$ 2.3} \\
    Md-Replay & 17.7 $\pm$ 4.3 & \textbf{42.9 $\pm$ 11.6} \\
    Random & 5.7 $\pm$ 3.1 & \textbf{19.4 $\pm$ 2.9} \\
 \bottomrule
\end{tabular}
\end{center}
\end{table}

\subsection{Teammate Modeling on SMAC Tasks}
\label{sec:app-vis-smac}
We show and analyze the quality of teammate modeling by \our-D on SMAC. 
Specifically, we choose two time steps from an episode on 3m map to analyze predictions on allies' attack targets and health points (HP), respectively.

\begin{figure}[h]
    \centering
    \begin{subfigure}[b]{0.46\linewidth}
        \includegraphics[width=\textwidth]{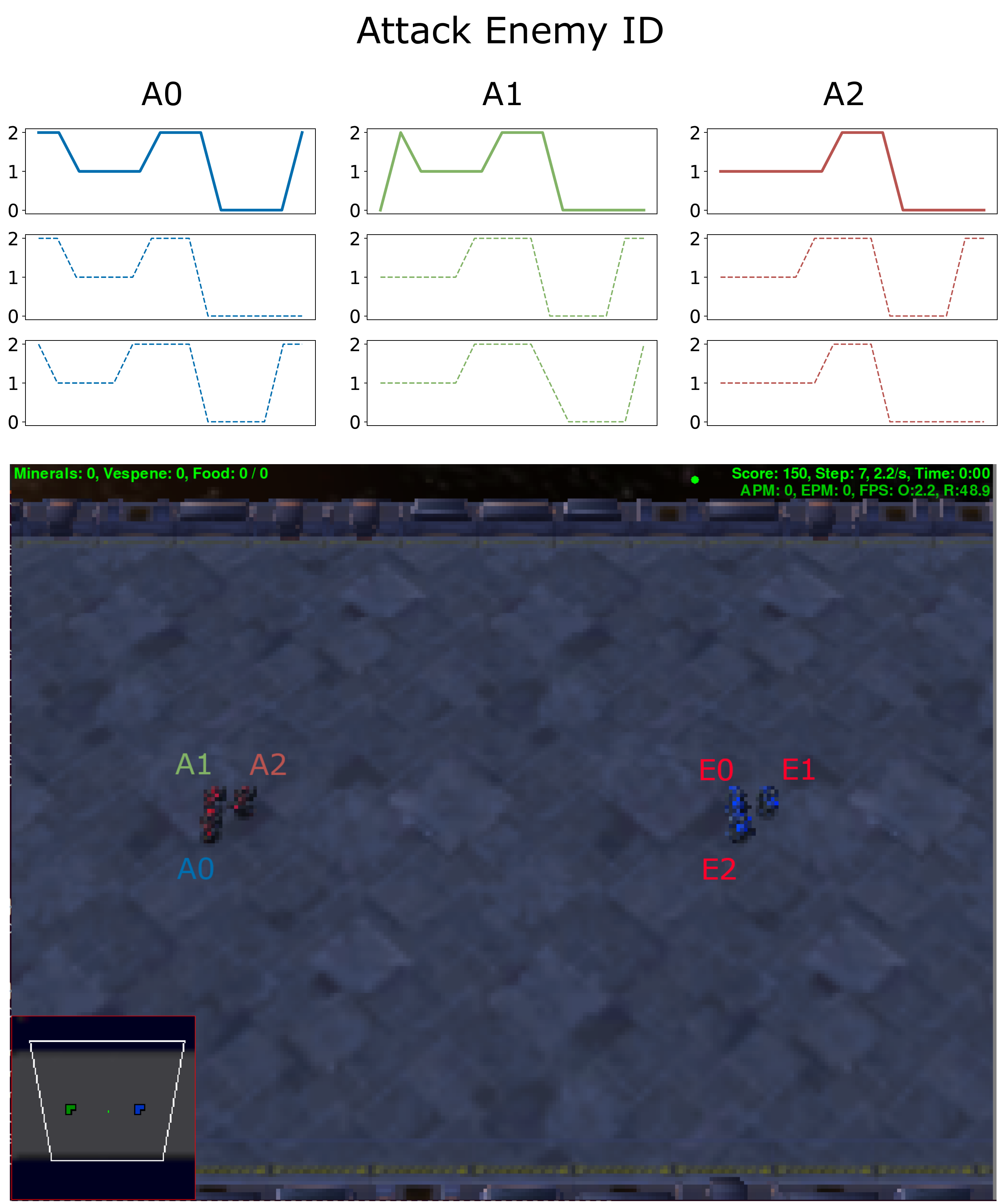}
        \caption{Ground-truth and predicted enemy's ID to attack by each ally agent.}
        \label{fig:oppo-attack}
    \end{subfigure}
    \hspace{5pt}
    \begin{subfigure}[b]{0.46\linewidth}
        \includegraphics[width=\textwidth]{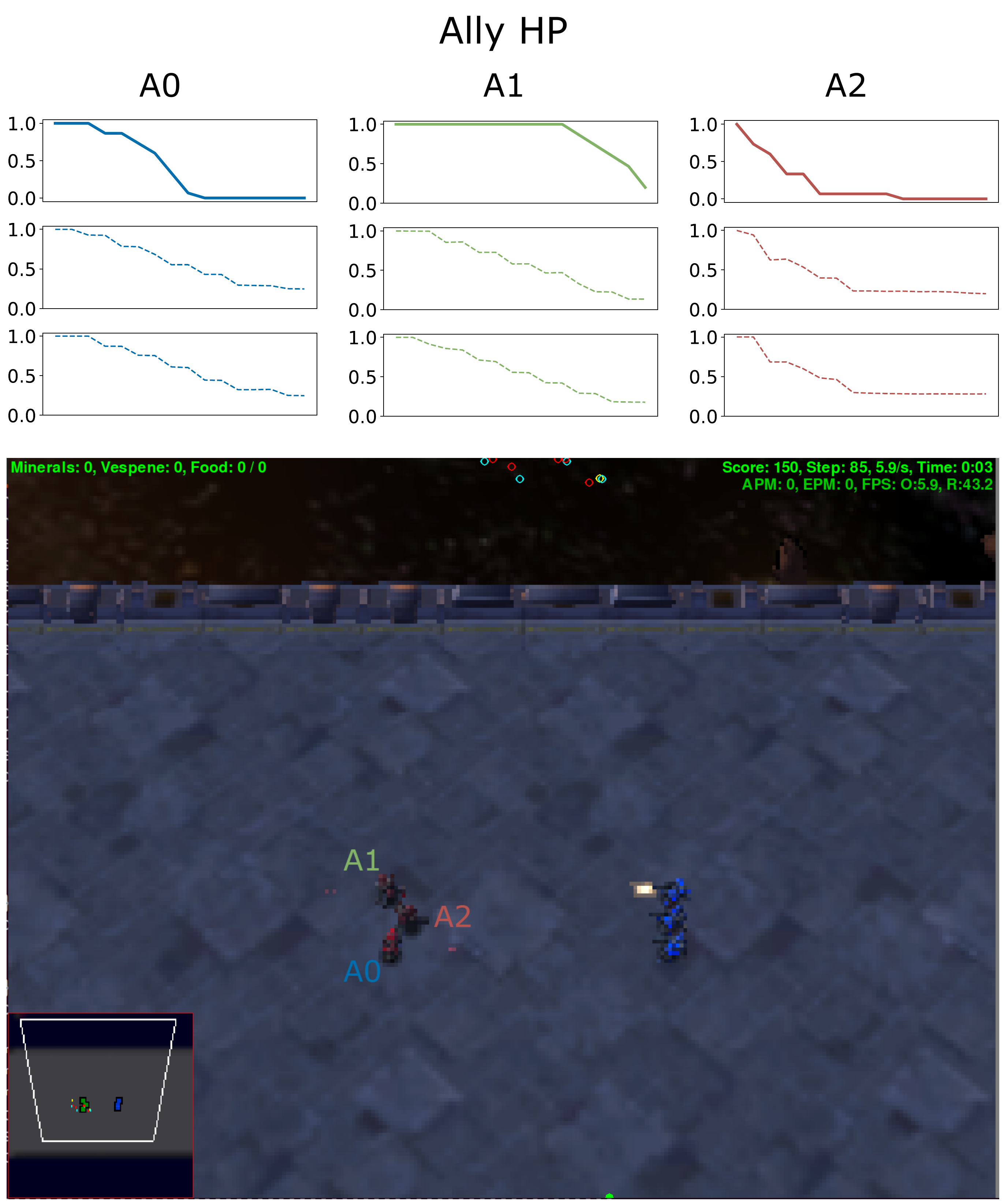}
        \caption{Ground-truth and predicted health points (HP) of each ally agent.}
        \label{fig:oppo-hp}
    \end{subfigure}
    \caption{The ground-truth and predicted information of different \our agents at two-time step. On the top of each figure, each column describes a different agent. The first row shows the change curve of the real value, and the last two rows below are the information predicted by other agents.}
\end{figure}

On top of \fig{fig:oppo-attack} is attacked enemy agent ID (0, 1, 2 stands for E0, E1, E2) of ally agents A0, A1, and A2. The first row is the ground-truth ID, and the second and the third rows are the predictions made by \our-D from the other two allies' views. We can see that the predictions are in general consistent with the ground-truth ID. As can be seen from the true values of the attack enemy ID, agents tend to focus their firepower on the same enemies at the same time. And the accurate prediction of allies' attack enemy IDs intuitively can help to execute such a strategy. 

In \fig{fig:oppo-hp}, we visualize the HP change curve of ally agents starting from another time step. From the environment state visualization below, agent A2 is the closest to enemies, so its HP drops the fastest. Such a pattern is successfully predicted by the other two agents. 

\subsection{Predicted Trajectory Visualization on NBA Dataset}
\label{sec:app-vis-nba}
\begin{figure}[h]
    \centering
    \includegraphics[width=\linewidth]{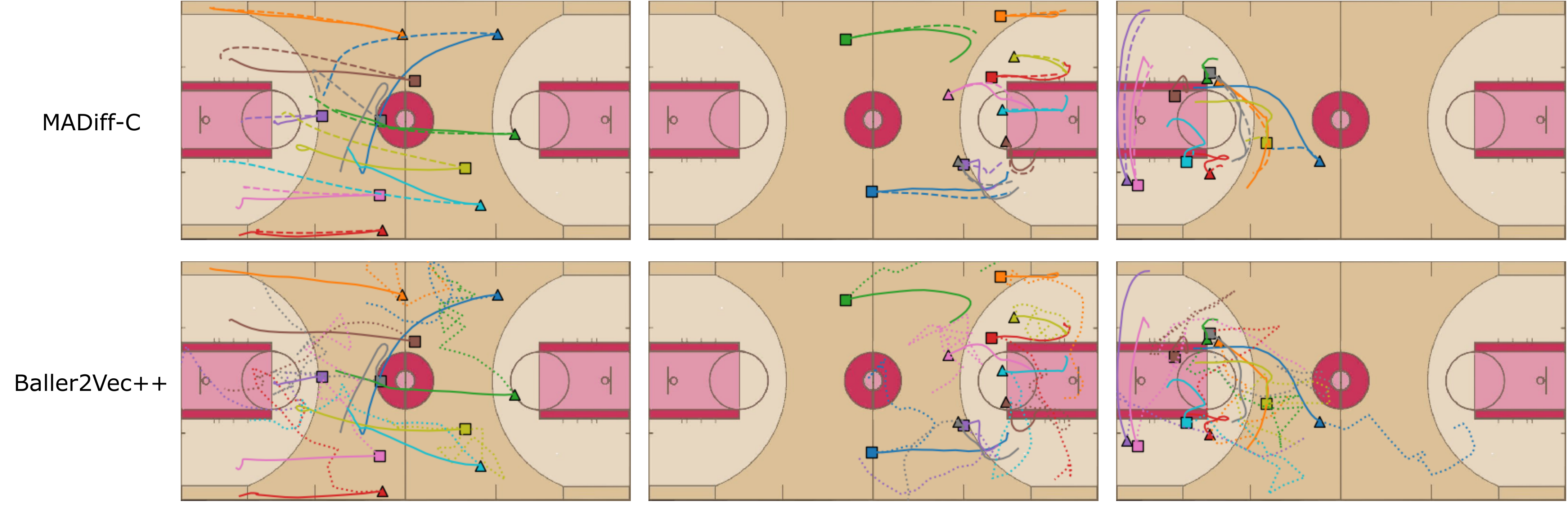}
    \caption{Real and Predicted multi-player trajectories by \our-C and Baller2Vec++.}
    \label{fig:nba-visualize}
\end{figure}

We visualize the players' moving trajectories predicted by \our-C and Baller2Vec++ on the NBA dataset in \fig{fig:nba-visualize}. In each image, the solid lines are real trajectories and the dashed lines are trajectories predicted by the model. The trajectories predicted by MADiff-C are closer to the real trajectories and are overall smoother compared to the Baller2Vec++ predictions.

\clearpage
\newpage
\section*{NeurIPS Paper Checklist}

\begin{enumerate}

\item {\bf Claims}
    \item[] Question: Do the main claims made in the abstract and introduction accurately reflect the paper's contributions and scope?
    \item[] Answer: \answerYes{} %
    \item[] Justification: The claims made in the abstract and introduction accurately reflect our contributions.

\item {\bf Limitations}
    \item[] Question: Does the paper discuss the limitations of the work performed by the authors?
    \item[] Answer: \answerYes{} %
    \item[] Justification: We point out the limitations of our method in \se{sec:limitation}.

\item {\bf Theory Assumptions and Proofs}
    \item[] Question: For each theoretical result, does the paper provide the full set of assumptions and a complete (and correct) proof?
    \item[] Answer: \answerNA{} %
    \item[] Justification: Our paper does not include theoretical results.

    \item {\bf Experimental Result Reproducibility}
    \item[] Question: Does the paper fully disclose all the information needed to reproduce the main experimental results of the paper to the extent that it affects the main claims and/or conclusions of the paper (regardless of whether the code and data are provided or not)?
    \item[] Answer: \answerYes{} %
    \item[] Justification: We illustrate our model architecture in \fig{fig:madiff} and list important hyperparameters in Appendix \se{sec:ap-hyperparameter}. We also provide the source code in supplementary materials.

\item {\bf Open access to data and code}
    \item[] Question: Does the paper provide open access to the data and code, with sufficient instructions to faithfully reproduce the main experimental results, as described in supplemental material?
    \item[] Answer: \answerYes{} %
    \item[] Justification: We provide code, anonymous data download link, and necessary instructions in supplementary materials.

\item {\bf Experimental Setting/Details}
    \item[] Question: Does the paper specify all the training and test details (e.g., data splits, hyperparameters, how they were chosen, type of optimizer, etc.) necessary to understand the results?
    \item[] Answer: \answerYes{} %
    \item[] Justification: We provide the experimental details in \se{sec:exp-task}, \se{sec:exp-baseline}, and Appendix \se{sec:appendix-addinfo-dataset}, \se{sec:appendix-baseline-implementation}, \se{ap:implementation}.

\item {\bf Experiment Statistical Significance}
    \item[] Question: Does the paper report error bars suitably and correctly defined or other appropriate information about the statistical significance of the experiments?
    \item[] Answer: \answerYes{} %
    \item[] Justification: We provide error bars in \tb{tb:marl-result}, \tb{tb:nba}, and \fig{fig:ablation-mpe}. Reported error bar is standard deviation calculated over trials with different random seeds.

\item {\bf Experiments Compute Resources}
    \item[] Question: For each experiment, does the paper provide sufficient information on the computer resources (type of compute workers, memory, time of execution) needed to reproduce the experiments?
    \item[] Answer: \answerYes{} %
    \item[] Justification: We provide concrete instances of both training and sampling wall time and resources of our algorithm in Appendix \se{ap-walltime}.

\item {\bf Code Of Ethics}
    \item[] Question: Does the research conducted in the paper conform, in every respect, with the NeurIPS Code of Ethics \url{https://neurips.cc/public/EthicsGuidelines}?
    \item[] Answer: \answerYes{} %
    \item[] Justification: Our paper conform with the NeurIPS Code of Ethics.

\item {\bf Broader Impacts}
    \item[] Question: Does the paper discuss both potential positive societal impacts and negative societal impacts of the work performed?
    \item[] Answer: \answerNA{} %
    \item[] Justification: The proposed algorithm is a general solution for a wide range of offline multi-agent learning problems. In our opinion, there is no specific societal impact that should be stated explicitly. 
    
\item {\bf Safeguards}
    \item[] Question: Does the paper describe safeguards that have been put in place for responsible release of data or models that have a high risk for misuse (e.g., pretrained language models, image generators, or scraped datasets)?
    \item[] Answer: \answerNA{} %
    \item[] Justification: Our paper poses no such risks.

\item {\bf Licenses for existing assets}
    \item[] Question: Are the creators or original owners of assets (e.g., code, data, models), used in the paper, properly credited and are the license and terms of use explicitly mentioned and properly respected?
    \item[] Answer: \answerYes{} %
    \item[] Justification: We cited and mentioned open-sourced implementation we used in Appendix \se{sec:appendix-baseline-implementation}.

\item {\bf New Assets}
    \item[] Question: Are new assets introduced in the paper well documented and is the documentation provided alongside the assets?
    \item[] Answer: \answerNA{} %
    \item[] Justification: Our paper does not publicly release new assets.

\item {\bf Crowdsourcing and Research with Human Subjects}
    \item[] Question: For crowdsourcing experiments and research with human subjects, does the paper include the full text of instructions given to participants and screenshots, if applicable, as well as details about compensation (if any)? 
    \item[] Answer: \answerNA{} %
    \item[] Justification: Our paper does not involve human subjects.

\item {\bf Institutional Review Board (IRB) Approvals or Equivalent for Research with Human Subjects}
    \item[] Question: Does the paper describe potential risks incurred by study participants, whether such risks were disclosed to the subjects, and whether Institutional Review Board (IRB) approvals (or an equivalent approval/review based on the requirements of your country or institution) were obtained?
    \item[] Answer: \answerNA{} %
    \item[] Justification: Our paper does not involve human subjects.

\end{enumerate}

\end{document}